\documentclass{article}
\usepackage{amssymb}
\usepackage{amsmath}
\usepackage{graphicx}
\usepackage{xspace}
\newcommand{\ourmodel}{DiWA\xspace}
\newcommand{\figref}[1]{Figure~\ref{#1}}
\newcommand{\tabref}[1]{Table~\ref{#1}}
\newcommand{\odrawer}{\texttt{\normalsize\fontseries{m}\selectfont open-drawer}}
\newcommand{\cdrawer}{\texttt{\normalsize\fontseries{m}\selectfont close-drawer}}
\newcommand{\lslider}{\texttt{\normalsize \fontseries{m}\selectfont move-slider-left}}
\newcommand{\rslider}{\texttt{\normalsize \fontseries{m}\selectfont move-slider-right}}
\newcommand{\lighton}{\texttt{\normalsize \fontseries{m}\selectfont turn-on-lightbulb}}
\newcommand{\lightoff}{\texttt{\normalsize \fontseries{m}\selectfont turn-off-lightbulb}}
\newcommand{\ledon}{\texttt{\normalsize \fontseries{m}\selectfont turn-on-LED}}
\newcommand{\ledoff}{\texttt{\normalsize \fontseries{m}\selectfont turn-off-LED}}

\newcommand\nnfootnote[1]{%
  \begin{NoHyper}
  \renewcommand\thefootnote{}\footnote{#1}%
  \addtocounter{footnote}{-1}%
  \end{NoHyper}
}

\newcommand{\liberoodrawer}{\texttt{\normalsize\fontseries{m}\selectfont open-top-drawer}}
\newcommand{\liberostove}{\texttt{\normalsize\fontseries{m}\selectfont turn-on-stove}}
\newcommand{\liberocbdrawer} {\texttt{\normalsize\fontseries{m}\selectfont close-bottom-drawer}}
\newcommand{\liberoctdrawer}{\texttt{\normalsize\fontseries{m}\selectfont close-top-drawer}}

\usepackage{multirow}
\usepackage{xcolor}
\usepackage{array}
\usepackage{booktabs}
\usepackage{microtype}
\usepackage{subfig}
\usepackage[font=footnotesize]{caption}
\usepackage{algorithm}
\usepackage{algpseudocode}
\usepackage{lscape}

\usepackage{hyperref}
\usepackage{soul}

\usepackage{mathtools}
\DeclarePairedDelimiterX{\infdivx}[2]{(}{)}{%
  #1\;\delimsize\|\;#2%
}

\definecolor{dred}{rgb}{1,0,0}

\usepackage[final]{corl_2025} % Uncomment for the camera-ready ``final'' version.
\title{DiWA: Diffusion Policy Adaptation with World Models}

\author{
  Akshay L Chandra$^{1*}$, Iman Nematollahi$^{1*}$, Chenguang Huang$^2$ \\
  \textbf{Tim Welschehold}$^1$\textbf{,} \textbf{Wolfram Burgard}$^2$\textbf{,} \textbf{Abhinav Valada}$^1$ \\
  $^1$ University of Freiburg \qquad $^2$ University of Technology Nuremberg\\[1em]
  \large\textbf{\url{https://diwa.cs.uni-freiburg.de}}\\[-1em]
}

\begin{document}
\maketitle\nnfootnote{$^*$ Equal contribution.}

%===============================================================================
\vspace{-0.8cm}
\begin{abstract}
    % We present \ourmodel, a framework for fine-tuning diffusion-based robotic skills with reinforcement learning entirely offline using a learned world model. \ourmodel adapts diffusion policies by practicing them over long-horizon rollouts in the latent space of a world model trained on unstructured play data, enabling skill refinement entirely offline using reinforcement learning. Compared to model-free baselines that require several million steps of physical interactions to fine-tune each individual skill, \ourmodel achieves effective adaptation using a world model trained once on only a few hundred thousand steps of unstructured play data, making it significantly more sample-efficient, and far more practical and safe for real-world robotics applications. On the challenging CALVIN benchmark, \ourmodel improves skill performance across 8 tasks using only offline adaptation, while requiring orders of magnitude fewer physical interactions than a model-free baseline. To the best of our knowledge, we are the first to demonstrate fine-tuning of diffusion policies for real-world robot skills within an offline world model.
    Fine-tuning diffusion policies with reinforcement learning (RL) presents significant challenges. The long denoising sequence for each action prediction impedes effective reward propagation.
    % Additionally, fine-tuning with standard RL methods is prohibitively expensive, often requiring millions of physical interactions. Prior work models the denoising steps in diffusion policies as a Markov Decision Process to adapt to RL policy updates, but its heavy reliance on environment interactions still leads to inefficiency.
    Moreover, standard RL methods require millions of real-world interactions, posing a major bottleneck for practical fine-tuning. Although prior work frames the denoising process in diffusion policies as a Markov Decision Process to enable RL-based updates, its strong dependence on environment interaction remains highly inefficient.
    To bridge this gap, we introduce \ourmodel, a novel framework that leverages a world model for fine-tuning diffusion-based robotic skills entirely offline with reinforcement learning. Unlike model-free approaches that require millions of environment interactions to fine-tune a repertoire of robot skills, \ourmodel achieves effective adaptation using a world model trained once on a few hundred thousand offline play interactions. This results in dramatically improved sample efficiency, making the approach significantly more practical and safer for real-world robot learning. On the challenging CALVIN benchmark, \ourmodel improves performance across eight tasks using only offline adaptation, while requiring orders of magnitude fewer physical interactions than model-free baselines. To our knowledge, this is the first demonstration of fine-tuning diffusion policies for real-world robotic skills using an offline world model.%We make the code publicly available at \url{https://diwa.cs.uni-freiburg.de}.
\end{abstract}
% Two or three meaningful keywords should be added here
\keywords{World Models, Imitation Learning, Reinforcement Learning} 

%===============================================================================

\section{Introduction}

Diffusion models have emerged as a powerful tool for robot policy learning, representing actions through conditional denoising processes that capture complex multi-modal behaviors~\citep{chi2023diffusion}. Their success stems from strong training stability and the ability to model high-dimensional distributions~\citep{ho2020denoising}. However, when trained purely through imitation learning on offline demonstrations, diffusion policies inherit the core limitations of imitation learning~\citep{ross2010efficient}, often struggle with distribution shifts, and fail in unseen scenarios due to imperfect or narrowly scoped expert trajectories. 
Reinforcement learning (RL) provides a natural path to overcome the limitations of imitation learning by enabling agents to improve through trial and error and explore beyond the constraints of the demonstration data. RL offers a general mechanism for fine-tuning pre-trained policies, allowing them to correct errors~\citep{nair2020awac,schmalstieg2022learning, nematollahi2022robot,mark2024policy,honerkamp2023n}, adapt to new situations~\citep{hu2024flare,nematollahi2023robot,luo2024precise}, and discover improved strategies~\citep{yang2024robot}. This pretrain-and-finetune paradigm, widely adopted in foundation models for language~\citep{brown2020language,ouyang2022training} and vision~\cite{radford2021learning,ruiz2023dreambooth}, is increasingly relevant in robotics. However, unlike those domains, fine-tuning in robotics demands physical interaction, making it significantly more challenging due to the sample inefficiency and safety concerns associated with deploying RL algorithms in the real world.

 \begin{figure}[t]
    \centering
    \includegraphics[width=0.9\columnwidth]{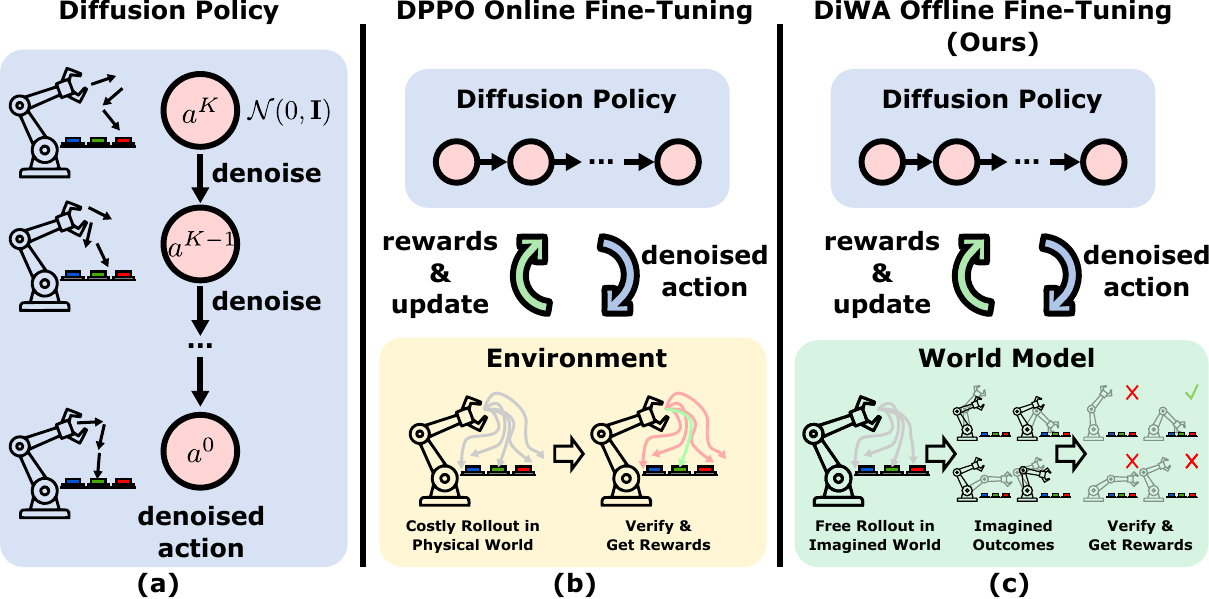}
    % \vspace{-0.6cm}
    \caption{(a) Standard diffusion policies trained via imitation learning are limited by offline data. (b) DPPO~\cite{ren2024diffusion} fine-tunes diffusion policies using online interactions, which are expensive and require access to real or simulated environments. (c) \textbf{DiWA} fine-tunes diffusion policies entirely offline through imagined rollouts in a learned world model, enabling safe and efficient policy improvement without additional physical interaction. \looseness=-1}
    \label{fig:motivation}
    \vspace{-0.6cm}
\end{figure}

A recent state-of-the-art method for fine-tuning diffusion policies is Diffusion Policy Policy Optimization (DPPO)~\citep{ren2024diffusion}, which uses Proximal Policy Optimization (PPO)~\citep{schulman2017proximal} to improve pre-trained diffusion models through on-policy reinforcement learning. DPPO shows that diffusion policies can be effectively fine-tuned with policy gradients, achieving strong results in simulation. However, it suffers from poor sample efficiency, requiring millions of interactions, which makes it impractical for real-world deployment where executing interactions is expensive, slow, and potentially unsafe. Although DPPO demonstrates zero-shot sim-to-real transfer, it relies on access to ground-truth state information from a high-fidelity simulator. These requirements make the direct application of DPPO-style online fine-tuning impractical for adapting robot skills in the real world, as the lack of low-level observations and the sim-to-real gap~\citep{chebotar2019closing} hinder reliable transfer from simulation. 
In contrast, humans can adapt their behavior with minimal physical trial-and-error by leveraging internal world models and an intuitive understanding of physics to anticipate outcomes and plan actions~\citep{matsuo2022deep}. Inspired by this ability, learned world models~\citep{ha2018world} have emerged as a powerful alternative to handcrafted simulators, enabling agents to improve policies through imagined interactions instead of costly online trials. These models compress high-dimensional observations into latent spaces that capture environment dynamics, allowing for long-horizon, on-policy rollouts in imagination~\citep{hafner2019dream,dreamerv2}. Recent work~\citep{nematolli2024lumos} demonstrates that language-conditioned policies trained purely within a world model can generalize to the real world without any additional physical fine-tuning, highlighting world models as a promising direction for safe and sample-efficient robot learning.

To enable sample-efficient and real-world compatible fine-tuning of diffusion policies, we introduce \textbf{\ourmodel}, a fully offline framework that leverages a learned world model instead of real or simulated environment interactions (see ~\figref{fig:motivation}). \ourmodel treats the world model as a safe, data-driven simulator, generating long-horizon imagined rollouts in latent space to fine-tune a pre-trained diffusion policy using on-policy reinforcement learning. This enables policy improvement through imagined practice in a learned ``dream'' of the environment, grounded in real data dynamics. By combining the expressiveness of diffusion models, the stability of policy gradients, and the imagination capabilities of learned world models, \ourmodel offers a practical and scalable approach for adapting robot skills without costly trial-and-error in the real world.

In summary, our contributions are threefold: 1) \textbf{Offline Fine-Tuning of Diffusion Policies via World Models:} We introduce \ourmodel, the first framework that fine-tunes diffusion policies entirely offline by leveraging a learned world model. By formulating a \emph{Dream Diffusion Markov Decision Process} (MDP), \ourmodel enables policy updates without any real or simulated interaction. 2) \textbf{Sample-Efficient Robot Skill Adaptation:} \ourmodel trains on unstructured play data to learn a latent world model and refines complex behaviors through imagined rollouts. It achieves significantly higher sample efficiency than baselines on the CALVIN benchmark. 3) \textbf{Zero-Shot Real-World Deployment:} We show that diffusion skills fine-tuned entirely within a learned world model trained on real-world play data can be deployed on real robots without requiring any additional physical interaction, enabling safe and effective real-world adaptation.\looseness=-1
\section{Related Work}
\label{sec:related_work}

\noindent\textbf{Reinforcement Learning for Robot Policy Adaptation:}
Imitation learning (IL) provides a sample-efficient way to train policies but often suffers from covariate shift and compounding errors when encountering out-of-distribution states. In contrast, Reinforcement Learning (RL) enables policy improvement through interaction with the environment, using reward signals to guide behavior. Since the success of deep Q-networks (DQN) on Atari~\citep{mnih2013playing}, RL has been widely adopted in robotics for tasks ranging from locomotion to manipulation~\citep{talpaert2019exploring, hwangbo2019learning, tai2017virtual}. A common paradigm combines IL and RL, first pre-training a base policy from demonstrations and then fine-tuning it using either online interactions~\citep{nematollahi2022robot, booher2024cimrl, lu2023imitation, rajeswaran2017learning, johannink2019residual,chen2025conrft} or reward signals extracted from offline data~\citep{vecerik2017leveraging, kumar2020conservative}. In this work, \ourmodel{} extends this two-stage framework to diffusion policies, enabling fine-tuning of pre-trained policies entirely offline via a learned world model.

\noindent\textbf{Reinforcement Learning with World Models:}
Due to the high cost and complexity of physical interactions in robotics, world models have emerged as a promising alternative for enabling sample-efficient reinforcement learning. These models~\citep{ha2018world} are predictive representations of environment dynamics that allow agents to plan and learn through imagined trajectories, reducing the need for real-world interaction. World models have been used for both (i) planning~\citep{NEURIPS2019_5faf461e, hafner2019learning, tdmpc,nematollahi22iros} and (ii) model-based rollouts to train policies~\citep{hafner2019dream, dreamerv2, dreamerv3}. However, most existing approaches operate in a closed-loop online setting, where the model is continuously updated using data collected by the learning agent, thereby tightly coupling the world model to the downstream task. An alternative paradigm is to learn general-purpose, task-agnostic world models from unstructured, unlabeled data such as play~\citep{demoss2023ditto, nematolli2024lumos}. These models can be reused across tasks by providing auxiliary reward signals or simulating interactions. \ourmodel follows this paradigm: it learns a general world model once from offline play data, freezes it, and uses it to fine-tune pre-trained policies entirely offline without any model updates.\looseness=-1

\noindent\textbf{Reinforcement Learning for Diffusion-Based Policies:}
Diffusion-based policies (DPs) have recently achieved strong performance in robotic imitation learning due to their stable training and capacity to model multi-modal behaviors \citep{chi2023diffusion, Chi2024UniversalMI, goyal2023rvt, sridhar2024nomad, pmlr-v229-xian23a, hou2024diffusion, team2024octo}. However, their effectiveness is constrained by the coverage and quality of expert demonstrations. To address this, several approaches have explored extending DPs with trajectory diffusion \citep{chen2024diffusion, ajay2022conditional, janner2022planning}, offline Q-learning~\citep{chen2022offline, ding2023consistency, wang2022diffusion}, online reinforcement learning~\citep{hansen2023idql, psenka2023learning, yang2023policy}, and residual learning~\citep{yuan2024policy}. Policy gradient methods~\citep{rl-book, Sutton1999PolicyGM}, which directly optimize the expected return of a policy, have also been applied to fine-tune diffusion models. This includes recent work on fine-tuning text-to-image diffusion models~\citep{NEURIPS2023_fc65fab8, wallace2024diffusion}, where the denoising process is treated as a multi-step MDP~\citep{black2023training, psenka2023learning, ren2024diffusion}. Our work builds directly on Diffusion Policy Policy Optimization (DPPO) \citep{ren2024diffusion}, which first demonstrated how to embed the diffusion denoising process into the environment MDP and apply PPO \citep{schulman2017proximal} for fine-tuning in control settings. While DPPO enables effective fine-tuning, it relies on online interactions and ground-truth environment signals. \ourmodel addresses this limitation by replacing the environment MDP with a learned world model, enabling offline fine-tuning entirely through imagined rollouts.\looseness=-1

\vspace{-0.5em}
\section{Problem Formulation}
\label{sec:problem}
\vspace{-0.5em}

We investigate the problem of offline fine-tuning of diffusion policies for robotic skill adaptation. We assume access to two types of offline datasets: a small set of expert demonstrations $\mathcal{D}_{\text{exp}}$ that are specific to the target skill, and a larger task-agnostic dataset of unstructured and unlabeled play $\mathcal{D}_{\text{play}}$. We model the real environment as a partially observable Markov Decision Process $\mathcal{M}_{\text{env}} = (\mathcal{S}, \mathcal{A}, P, R, \gamma)$, where $\mathcal{S}$ is the state-observation space, $\mathcal{A}$ the continuous action space, $P(s_{t+1} \mid s_t, a_t)$ the transition dynamics, $R(s_t,a_t)$ the reward function, and $\gamma \in (0,1)$ the discount factor. A diffusion policy $\pi_\theta(a_t \mid s_t)$ generates actions by first sampling Gaussian noise $\bar{a}_t^K \sim \mathcal{N}(0, I)$, then progressively denoising it through learned transitions:
\begin{equation}
\label{eq:diff_pi}
\bar{a}_t^{k-1} \sim \pi_\theta(\bar{a}_t^{k-1} \mid s_t, \bar{a}_t^{k}), \quad \text{for} \quad k = K, K-1, \dots, 1,
\end{equation}
where the final output $\bar{a}_t^{0}$ is taken as the environment action $a_t$. The diffusion policy $\pi_\theta$ is first pre-trained via behavior cloning on $\mathcal{D}_{\text{exp}}$, imitating expert actions through denoising. However, behavior cloning is limited by distribution shift and the quality of demonstrations. To address this, we fine-tune the pre-trained policy to maximize expected cumulative reward in the real environment:\looseness=-1
\begin{equation}
\theta^\star = \arg\max_\theta \mathbb{E}_{\tau \sim \pi_\theta} \left[ \sum_{t=0}^\infty \gamma^t R(s_t, a_t) \right].
\end{equation}
Direct fine-tuning in $\mathcal{M}_{\text{env}}$ is impractical due to high sample complexity and real-world safety concerns. Instead, we train a latent dynamics model on $\mathcal{D}_{\text{play}}$ and define a world model MDP $\mathcal{M}_{\text{wm}} = (\mathcal{Z}, \mathcal{A}, P_\phi, R_\psi, \gamma)$, where $\mathcal{Z}$ is the learned latent space. Fine-tuning is then performed entirely within $\mathcal{M}_{\text{wm}}$, allowing for efficient and safe offline policy adaptation through imagined rollouts.\looseness=-1
\vspace{-0.5em}
\section{Offline Adaptation of Diffusion Policy with DiWA}
\label{sec:method}
\begin{figure}[t]
    \centering
    \includegraphics[width=1.0\columnwidth]{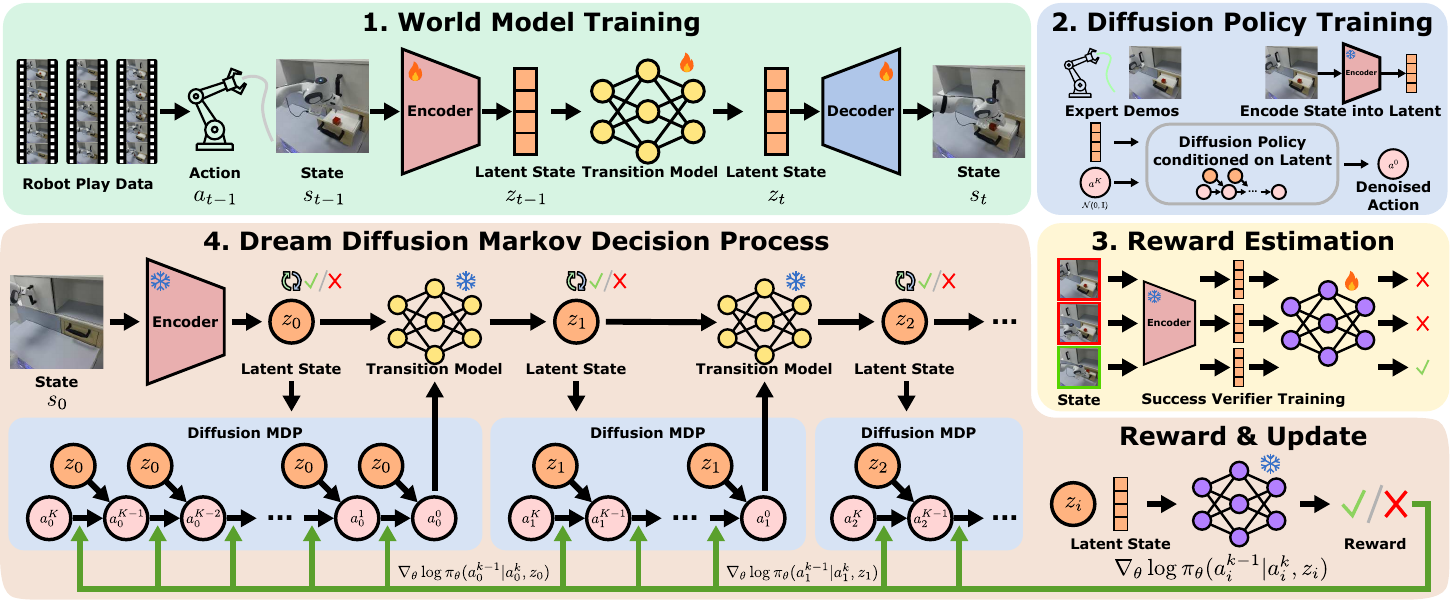}
    \caption{\textbf{\ourmodel} framework: (1) A world model is trained on unstructured robot play data to learn latent dynamics. (2) A diffusion policy is pre-trained on expert demonstrations using learned latent representations. (3) A success classifier is trained on expert rollouts to estimate task rewards. (4) The diffusion policy is fine-tuned entirely offline via imagined rollouts within the Dream Diffusion MDP, using policy gradients and classifier-based rewards.\looseness=-1}
    \label{fig:model_overview}
    \vspace{-0.3cm}
\end{figure}

\vspace{-0.5em}
In this section, we introduce \textbf{\ourmodel}. The training process consists of four phases: (1) learning a world model from an unlabeled play dataset $\mathcal{D}_{\text{play}}$, (2) pretraining a diffusion policy to imitate expert actions from latent representations of $\mathcal{D}_{\text{exp}}$, (3) training a reward classifier on those latents to equip the world model with a task-specific reward, and (4) fine-tuning the policy entirely within the latent space of the world model. At inference time, the fine-tuned policy is deployed in the real environment without any additional adaptation. \figref{fig:model_overview} provides an overview of the approach. For details on hyperparameters, architecture choices, and the pseudocode of \ourmodel please refer to Appendix~\ref{sec:hyperparams} and Algorithm~\ref{alg:diwa}.\looseness=-1
\vspace{-0.5em}
\subsection{World Model Learning}
\vspace{-0.5em}
We train a latent dynamics model on the unlabeled play dataset $\mathcal{D}_{\text{play}}$ to enable offline policy adaptation. The learned world model defines a latent-space MDP $\mathcal{M}_{\text{wm}} = (\mathcal{Z}, \mathcal{A}, P_\phi)$, where $\mathcal{Z}$ is the learned latent space and $P_\phi$ denotes the transition dynamics. Following prior work~\cite{dreamerv2, nematolli2024lumos}, we use a recurrent state-space model architecture with an encoder, dynamics model, and decoder. At each timestep $t$, the model maintains a deterministic recurrent state $h_t$ updated by a transition function $f_\phi$, and samples a stochastic latent variable $z_t$ from a posterior conditioned on the current observation $x_t$:\looseness=-1
\begin{equation}
\label{eq:model_components}
\begin{aligned}
    &\text{Recurrent state:}     &h_t     &= f_\phi(\hat{s}_{t-1}, a_{t-1}) \quad &\text{Representation model:} \quad z_t     &\sim q_\phi(z_t \mid h_t, x_t) \\
    &\text{Dynamics predictor:}  &\hat{z}_t &\sim p_\phi(\hat{z}_t \mid h_t) \quad &\text{Decoder:} \quad    \hat{x}_t &\sim p_\phi(\hat{x}_t \mid \hat{s}_t),
\end{aligned}
\end{equation}
where the model state is $\hat{s}_t = (h_t, z_t)$. The posterior $q_\phi$ and prior $p_\phi$ are modeled as categorical distributions, optimized using straight-through gradient estimators~\cite{bengio2013estimating}. The model parameters $\phi$ are trained by minimizing the negative variational evidence lower bound (ELBO):\looseness=-1
\begin{equation}
\label{eq:elbo}
\min_{\phi}\; \mathbb{E}_{q_\phi}\left[
\sum_{t=1}^T -\log p_\phi(x_t \mid s_t) + \beta \, \text{KL}\left(q_\phi(z_t \mid h_t, x_t)\, \| \, p_\phi(z_t \mid h_t)\right)
\right],
\end{equation}
where $\beta$ controls KL regularization. After training, the world model generates imagined trajectories by rolling out latent states from the learned prior $\hat{z}_t \sim p_\phi(\hat{z}_t \mid h_t)$ without additional observations.
\subsection{Pre-training Diffusion Policies}
We pre-train the diffusion policy via behavior cloning on expert demonstrations from $\mathcal{D}_{\text{exp}}$. Observations are encoded into latents using the world model, and the policy learns to iteratively denoise random noise into expert actions. This maximizes the likelihood of demonstrated behavior and provides the initialization for offline fine-tuning within the Dream Diffusion MDP.\looseness=-1

\subsection{Latent Reward Estimation from Expert Demonstrations}
The world model, trained on task-agnostic play data, lacks a reward signal aligned with the target skill. To address this, we train a binary classifier $C_\psi(z_t)$ on latent states extracted from expert demonstrations $\mathcal{D}_{\text{exp}}$. Each observation $x_t$ is encoded into a latent $z_t$ using the world model encoder, and the classifier is trained to predict task success by treating latents from annotated successful frames as positives. During imagined rollouts in $\mathcal{M}_{\text{wm}}$, rewards are computed as $R_\psi(z_t, a_t) := C_\psi(z_{t+1})$, where $C_\psi(z_{t+1}) \in [0,1]$ reflects the probability of success. This results in an augmented MDP ${\mathcal{M}}_{\text{wm}} = (\mathcal{Z}, \mathcal{A}, P_\phi, R_\psi, \gamma)$ that supports fully offline fine-tuning in imagined trajectories.\looseness=-1
\subsection{Dream Diffusion MDP}
As observed in prior work~\cite{black2023training, psenka2023learning, ren2024diffusion}, a diffusion denoising process can be represented as a multi-step MDP where the likelihood at each step is accessible. We extend this formalism by embedding the diffusion denoising process into the world model MDP, forming the \emph{Dream Diffusion MDP} $\mathcal{M}_{\text{DD}}$. Let $\bar{t}(t,k) = tK + (K-k)$ index the denoising steps across world model timesteps $t$ and denoising steps $k$, where $K$ is the total number of denoising steps and $k$ decreases lexicographically from $K$ to $1$. At index $\bar{t}(t,k)$, the Dream Diffusion MDP defines the state, action, and reward as\looseness=-1
\begin{equation}
\bar{s}_{\bar{t}(t,k)} = (z_t, \bar{a}_t^{k}), \quad
\bar{a}_{\bar{t}(t,k)} = \bar{a}_t^{k-1}, \quad
\bar{R}_{\bar{t}(t,k)} =
\begin{cases}
R_\psi(z_t, \bar{a}_t^0), & \text{if } k=1,\\
0, & \text{otherwise}.
\end{cases}
\end{equation}
Here, $\bar{a}_t^{k}$ denotes the intermediate action at denoising step $k$. The transition dynamics are given by\looseness=-1
\begin{equation}
\bar{P}(\bar{s}_{\bar{t}+1} \mid \bar{s}_{\bar{t}}, \bar{a}_{\bar{t}}) =
\begin{cases}
\delta(z_t, \bar{a}_t^{k-1}), & \text{if } k > 1,\\
P_\phi(z_{t+1} \mid z_t, \bar{a}_t^0) \otimes \mathcal{N}(0, I), & \text{if } k = 1,
\end{cases}
\end{equation}
where $\delta(\cdot)$ denotes a Dirac distribution. At denoising steps $k>1$, the diffusion policy iteratively denoises $\bar{a}_t^k$ into $\bar{a}_t^{k-1}$ while remaining at latent state $z_t$. When $k=1$, the final action $\bar{a}_t^0$ is produced, the world model transitions to $z_{t+1}$, and a new diffusion process begins from fresh noise. Following Eq.~\eqref{eq:diff_pi}, the policy at each inner step of the Dream Diffusion MDP is parameterized as a Gaussian:\looseness=-1
\begin{equation}
\bar{\pi}_\theta(\bar{a}_t^{k-1} \mid z_t, \bar{a}_t^{k}) = \mathcal{N}\left(\bar{a}_t^{k-1}; \mu_\theta(z_t, \bar{a}_t^k, k), \sigma_{k}^2 I\right),    
\end{equation}
where $\mu_\theta$ is a neural network output. Since each denoising step defines a Gaussian likelihood, the Dream Diffusion MDP admits a well-defined policy gradient objective. Specifically, we optimize
\begin{equation}
\nabla_\theta \bar{\mathcal{J}}(\bar{\pi}_\theta) = \mathbb{E}^{\bar{\pi}_\theta, \bar{P}}\left[\sum_{\bar{t} \geq 0} \nabla_\theta \log \bar{\pi}_\theta(\bar{a}_{\bar{t}} \mid \bar{s}_{\bar{t}}) \, \bar{r}(\bar{s}_{\bar{t}}, \bar{a}_{\bar{t}})\right],    
\end{equation}
where $\bar{r}(\bar{s}_{\bar{t}}, \bar{a}_{\bar{t}}) := \sum_{\tau \geq \bar{t}} \gamma^{\tau} \bar{R}(\bar{s}_\tau, \bar{a}_\tau)$ denotes the return. This objective corresponds to the expected cumulative reward over denoising steps and enables gradient-based fine-tuning of diffusion policies through rollouts in the imagined latent space.
\subsection{Fine-tuning within Dream Diffusion MDP}
We fine-tune the diffusion policy in the Dream Diffusion MDP $\mathcal{M}_{\text{DD}}$ using Proximal Policy Optimization (PPO)~\citep{schulman2017proximal}. Inspired by the two-layer structure of DPPO~\citep{ren2024diffusion}, we adapt PPO to operate entirely within imagined rollouts, alternating between denoising steps and latent transitions. The PPO objective is defined as
\begin{equation}
\mathcal{L}_\text{PPO} = \mathbb{E}^{\bar{\pi}_{\theta_\text{old}}}_{(\bar{s}, \bar{a})}\left[
\min\left(
\rho_\theta(\bar{s}, \bar{a}) \hat{A}(\bar{s}, \bar{a}),\ 
\operatorname{clip}(\rho_\theta(\bar{s}, \bar{a}), 1 - \epsilon, 1 + \epsilon) \hat{A}(\bar{s}, \bar{a})
\right)
\right],    
\end{equation}
where $\rho_\theta$ is the importance sampling ratio between the new and old policies. The clipping threshold $\epsilon$ constrains the policy update to ensure stability. We estimate the advantage at the denoising step $k$ as
\begin{equation}
\hat{A}(\bar{s}_{\bar{t}(t,k)}, \bar{a}_{\bar{t}(t,k)}) = \gamma_\text{denoise}^k \left( \bar{r}(\bar{s}_{\bar{t}}, \bar{a}_{\bar{t}}) - \hat{V}(z_t) \right),    
\end{equation}
where $\gamma_\text{denoise} \in (0,1)$ downweights the contribution of earlier, noisier denoising steps, and $\hat{V}$ estimates the value from the latent state $z_t$.

To enhance stability and ensure reliable transfer to the real environment, we augment the fine-tuning objective with a behavior cloning (BC) regularization term. Although world models trained on large play datasets capture environment dynamics well, they may still contain subtle errors that the RL agent can exploit, resulting in policies that perform well in imagination but fail in the real environment~\cite{modelexploit}. To address this, we constrain the updated policy to remain close to the pre-trained diffusion policy~\cite{torne2024reconciling}. The resulting objective is\looseness=-1
\begin{equation}
\mathcal{L}_\theta = \mathcal{L}_\text{PPO} - \alpha_\text{BC} \ \mathbb{E}^{\bar{\pi}_{\theta_\text{old}}} \left[\sum_{k=1}^{K} \log \pi_{\theta_\text{pre}}(\bar{a}_t^{k-1} \mid  z_t, \bar{a}_t^{k})\right],    
\end{equation}
where $\pi_{\theta_\text{pre}}$ is the frozen pre-trained policy and $\alpha_\text{BC}$ controls the strength of the regularization.\looseness=-1

\section{Experimental Evaluation}
\label{sec:result}
We evaluate \ourmodel{} for fine-tuning diffusion policies in both simulation and the real-world. Our goals are to: (i) assess whether \ourmodel{} can effectively fine-tune policies entirely offline and achieve high task success without additional environment interaction;  
(ii) analyze the impact of world model fidelity and reward classifier accuracy on adaptation performance; and  
(iii) evaluate the approach's ability to scale to real-world robotic tasks and transfer zero-shot from imagination to physical execution.

\subsection{Simulation Results}
\label{subsec:simulation_results}
We evaluate our method in environment \textit{D} of the CALVIN simulator~\citep{mees2022calvin}, which features a 7-DoF Franka Emika Panda robot performing diverse tabletop manipulation tasks. CALVIN offers a teleoperated play dataset that is both broad in coverage and easy to collect, making it ideal for training task-agnostic world models. We train the world model on six hours of play data ($\sim$500,000 transitions) and use a small annotated subset (50 demonstrations per skill) to pre-train individual diffusion policies.  Evaluation is conducted on eight tasks from the benchmark (for experiments on LIBERO benchmark~\citep{liu2023libero}, see Appendix~\ref{sec:libero}).\looseness=-1
\looseness=-1

\textbf{Evaluation Protocol:} \hspace{0.05cm}
We compare \ourmodel to Diffusion Policy Policy Optimization (DPPO)~\citep{ren2024diffusion}, which fine-tunes diffusion policies via PPO by treating the denoising process as a multi-step MDP. While DPPO baselines fine-tune policies through direct environment interactions (in simulation or real-world), \ourmodel performs fine-tuning entirely offline using imagined rollouts within the latent space of a learned world model. We evaluate DPPO in two variants. DPPO (Vision), the original variant introduced by Ren et al., takes raw pixel observations as input using a Vision Transformer (ViT) encoder~\cite{hu2023imitation}. DPPO (Vision WM Encoder) instead replaces the ViT with the same world-model encoder as \ourmodel, so that both methods start from identical pre-trained diffusion policies and receive the same latent state input for each skill. A key difference between the two settings lies in reward supervision: DPPO uses the ground-truth task completion signal available in the real environment, whereas \ourmodel relies on a learned reward classifier trained from a small set of expert demonstrations, introducing an additional challenge for policy optimization. For \ourmodel, we report the performance improvement achieved after 5 million fine-tuning steps conducted entirely in the latent space of the world model. For DPPO baselines, we measure the number of real environment interactions required to match the performance of \ourmodel.
\begin{table}[t]
\centering
\caption{\ourmodel successfully fine-tunes diffusion policies entirely offline using imagined rollouts in a learned world model. In contrast, DPPO requires hundreds of thousands of online interactions to achieve comparable performance. The DPPO (Vision) variant, operating directly on raw RGB observations without world‑model latents, requires far more interactions to reach similar performance. Results are averaged over three random seeds.\looseness=-1}
\resizebox{\textwidth}{!}{%
\begin{tabular}{cc|c|cc}
\midrule
\multicolumn{1}{c|}{\multirow{3}{*}{\textbf{Task}}} & \textbf{Base}             & \textbf{DiWA (Ours)}               & \multicolumn{1}{c|}{\textbf{\begin{tabular}[c]{@{}c@{}}DPPO \\ (Vision WM Encoder)\end{tabular}}} & \textbf{\begin{tabular}[c]{@{}c@{}}DPPO \\ (Vision)\end{tabular}} \\ \cmidrule{3-5} 
\multicolumn{1}{c|}{}          & \textbf{Diffusion Policy} & \textbf{Offline Fine-Tuning}       & \multicolumn{2}{c}{\textbf{Online Fine-Tuning}}                                                                                                                      \\ \cmidrule{2-5} 
\multicolumn{1}{c|}{}          & \textbf{Success Rate}     & \textbf{Success Rate}              & \multicolumn{2}{c}{\textbf{Env Steps to Match DiWA}}                                                                                                                 \\ \midrule
\multicolumn{1}{c|}{\odrawer}  & $57.78 \pm 3.85$   & $\textbf{74.44} \pm \textbf{1.92}$ & \multicolumn{1}{c|}{$117{,}600 \pm 23{,}758$}                                                     & $134{,}400 \pm 26{,}508$                                          \\
\multicolumn{1}{c|}{\cdrawer}  & $59.14 \pm 5.08$   & $\textbf{91.95} \pm \textbf{1.99}$ & \multicolumn{1}{c|}{$345{,}600 \pm 27{,}651$}                                                     & $1{,}545{,}600 \pm 261{,}346$                                          \\
\multicolumn{1}{c|}{\lslider}  & $62.15 \pm 0.60$   & $\textbf{83.33} \pm \textbf{1.80}$ & \multicolumn{1}{c|}{$270{,}933 \pm 28{,}780$}                                                     & $1{,}377{,}600 \pm 251{,}439$                                          \\
\multicolumn{1}{c|}{\rslider}  & $62.55 \pm 3.55$   & $\textbf{82.76} \pm \textbf{3.45}$ & \multicolumn{1}{c|}{$249{,}600 \pm 09{,}050$}                                                     & $537{,}600 \pm 23{,}758$                                          \\
\multicolumn{1}{c|}{\lighton}  & $60.61 \pm 3.03$   & $\textbf{91.92} \pm \textbf{1.75}$ & \multicolumn{1}{c|}{$302{,}933 \pm 15{,}964$}                                                     & $588{,}000 \pm 62{,}859$                                          \\
\multicolumn{1}{c|}{\lightoff} & $35.63 \pm 1.99$   & $\textbf{77.01} \pm \textbf{1.99}$ & \multicolumn{1}{c|}{$327{,}066 \pm 13{,}546$}                                                     & $1{,}260{,}000 \pm 142{,}552$                                          \\
\multicolumn{1}{c|}{\ledon}    & $48.43 \pm 3.67$   & $\textbf{86.21} \pm \textbf{3.45}$ & \multicolumn{1}{c|}{$494{,}933 \pm 45{,}655$}                                                     & $2{,}251{,}200 \pm 33{,}940$                                          \\
\multicolumn{1}{c|}{\ledoff}   & $55.25 \pm 4.79$   & $\textbf{82.33} \pm \textbf{6.53}$ & \multicolumn{1}{c|}{$277{,}333 \pm 31{,}928$}                                                     & $184{,}800 \pm 23{,}758$                                          \\ \midrule
\multicolumn{2}{c|}{\textbf{Total Physical Interactions:}}                      & \textbf{0}                         & \multicolumn{1}{c|}{$\sim$2.5M}                                                                   & $\sim$8M                                                        \\ \midrule
\end{tabular}%
}
\label{tab:sim_exp}
\vspace{-0.5cm}
\end{table}

\tabref{tab:sim_exp} reports the average success rates of pre-trained diffusion policies and their fine-tuned counterparts, averaged over three random seeds. \ourmodel successfully fine-tunes all evaluated robotic manipulation skills entirely offline, without requiring any additional physical interaction. In contrast, DPPO baselines typically require several hundred thousand environment interactions to reach a similar level of performance. Importantly, these interactions involve online exploration, which is often unsafe or impractical in real-world robotic settings. Overall, these results highlight that \ourmodel enables effective skill adaptation using only offline data, offering a safer and more sample-efficient alternative to model-free approaches. Among DPPO baselines, the world-model latent variant performs best, indicating that the latents learned by our world model are richer than ViT-based image encodings. See Appendix~\ref{sec:dppo_variants} for a more detailed comparison of DPPO variants.\looseness=-1
\begin{figure}[b]
    \vspace{-0.4cm}
    \centering
    \includegraphics[width=1.0\linewidth]{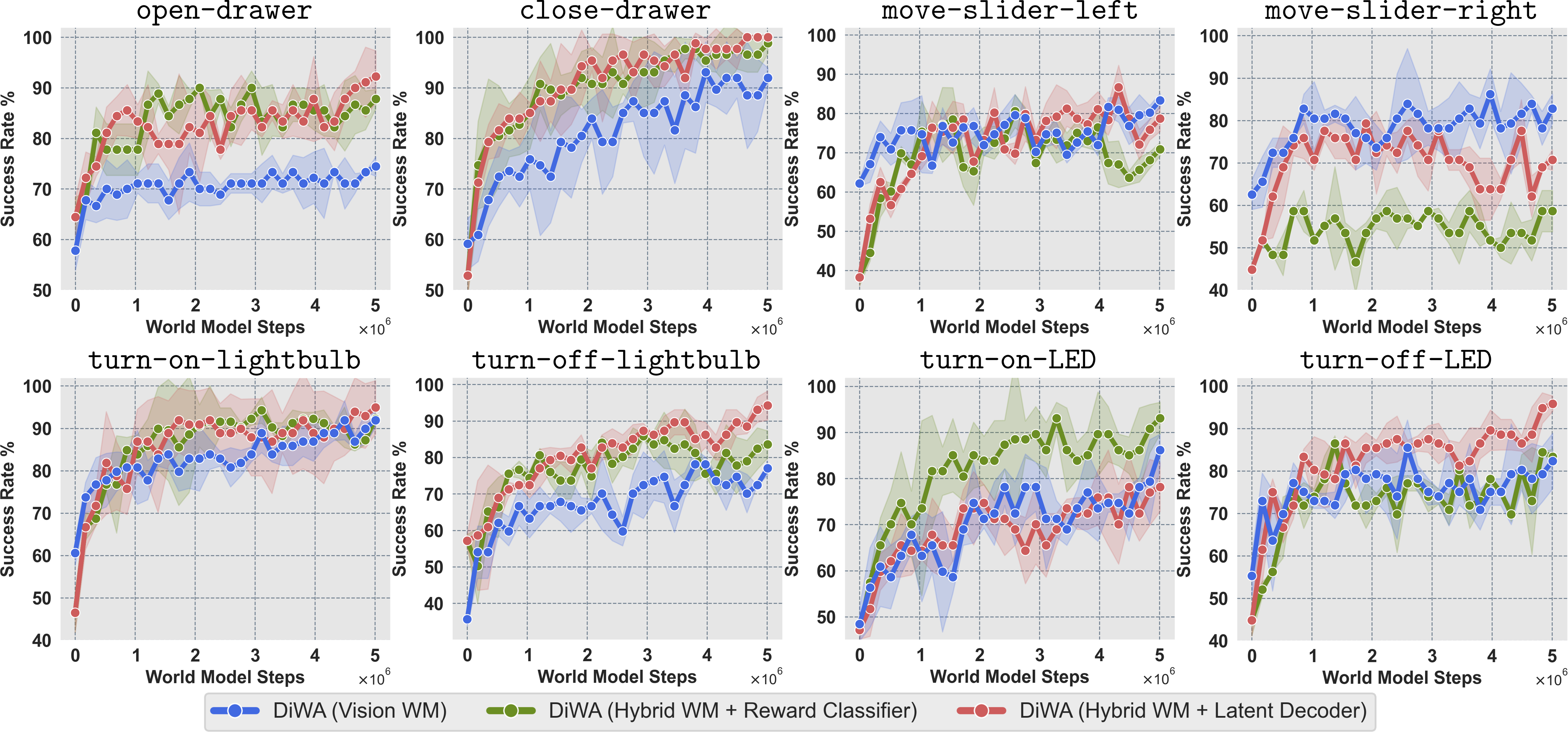}
    \caption{Comparison of three \ourmodel{} variants on simulated fine-tuning tasks. Blue uses only visual inputs, while green and red both incorporate scene state supervision. Red further decodes rewards from latents instead of relying on a learned classifier. Results demonstrate that more expressive world models and more accurate reward signals lead to improved offline fine-tuning performance.}
    \label{fig:visionwm-vs-hybridwm}
    \vspace{-0.4cm}
\end{figure}

To assess the impact of model components on fine-tuning performance, we compare three variants of our model: (i) \ourmodel (Vision WM), which uses a world model trained only on visual observations; (ii) \ourmodel (Hybrid WM + Reward Classifier), which incorporates both visual inputs and privileged scene state during training but still relies on a learned reward classifier; and (iii) \ourmodel (Hybrid WM + Latent Decoder), which also uses scene-state-conditioned latents but infers rewards by decoding them into scene state and applying a reward function directly. ~\figref{fig:visionwm-vs-hybridwm} highlights the differences across these model variants.
Comparing the first two variants, we find that hybrid world models enable faster and more stable fine-tuning, likely due to more accurate latent dynamics learned from scene state supervision, which improves the quality of imagined rollouts. Next, comparing the two hybrid variants, we isolate the effect of the reward function: latent decoding leverages scene-aware latents to reconstruct state variables, which enables more reliable reward computation and often yields stronger fine-tuning performance (see Appendix~\ref{sec:reward_clf} for a precision–recall analysis of our reward classifier). While we focus on \ourmodel (Vision WM) as our main variant because it relies solely on visual inputs and is thus compatible with real-world robotic setups, these results indicate that more expressive world models and more accurate reward signals can substantially enhance fine-tuning performance.\looseness=-1

\subsection{Real-World Results}
\begin{figure*}[t]
    \centering
    \subfloat[Real-world Manipulation Skills]{
        \includegraphics[height=4.cm]{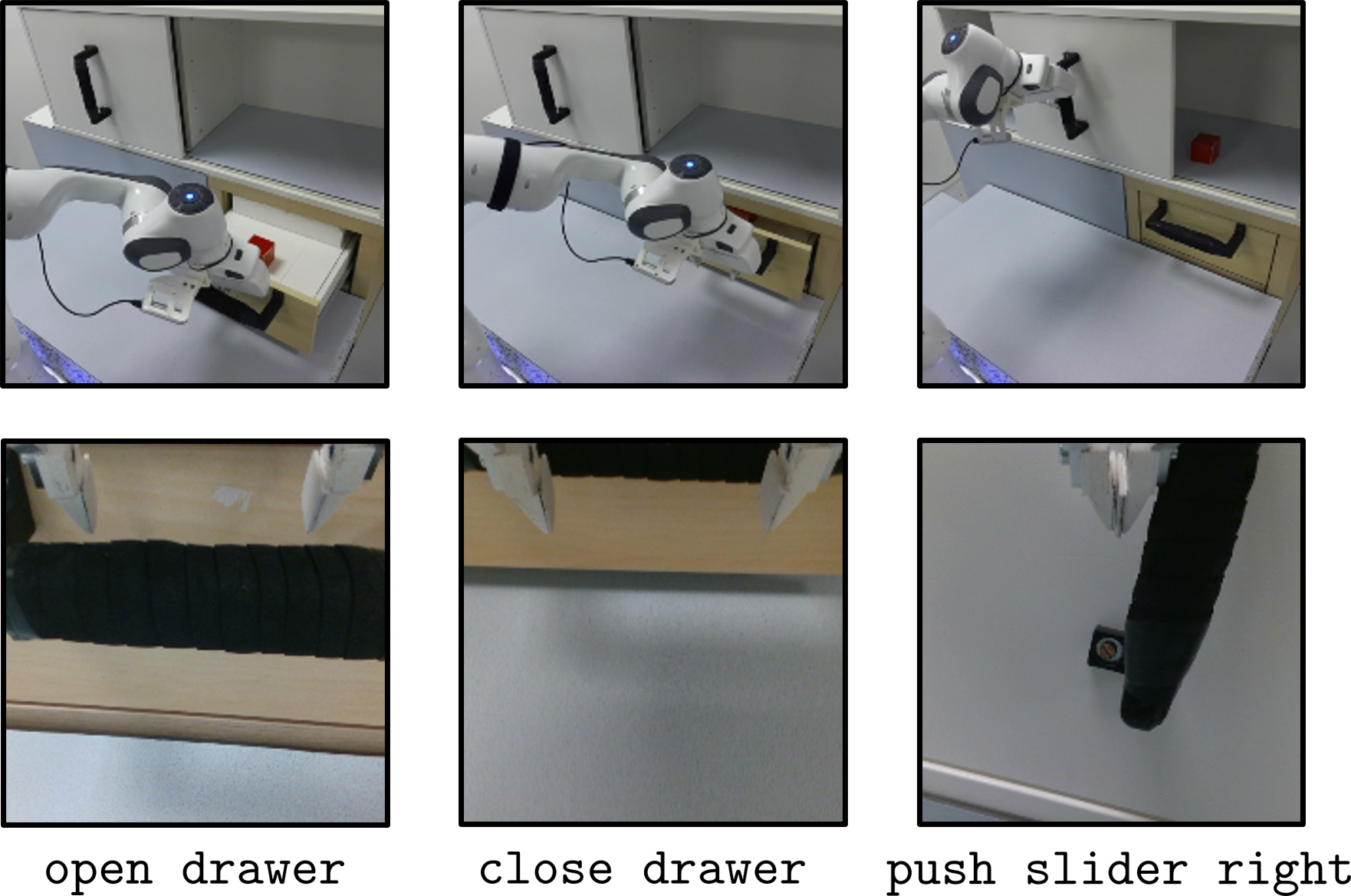}\label{fig:real_skills}}
    \hspace{1cm}
    \subfloat[Real-World Fine-Tuning Results]{
        \includegraphics[height=4.cm]{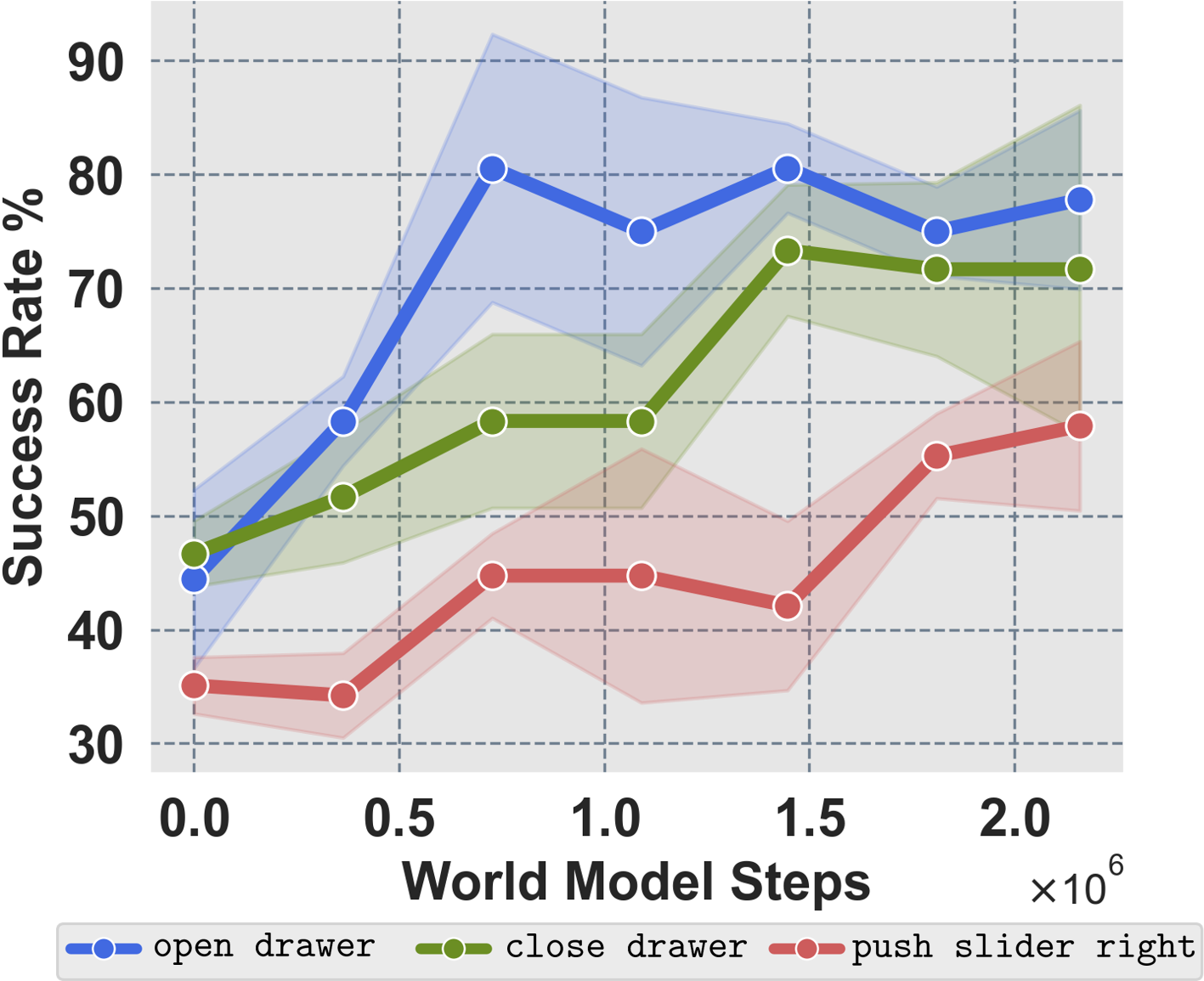}\label{fig:real_adapt}}
        \caption{(a) The three real-world manipulation tasks used for evaluation. (b) Success rates before and after offline fine-tuning with \ourmodel, averaged over 20 rollouts and three seeds. Values correspond to checkpoints saved during fine-tuning. While pre-trained diffusion policies show limited initial performance, \ourmodel enables significant improvement through imagination-based reinforcement learning without physical interaction.}
    \label{fig:real_plots}
    \vspace{-0.5cm}
\end{figure*}

To evaluate \ourmodel on real-world robotic skills, we conducted experiments with a Franka Emika Panda robot operating in a tabletop environment containing a cabinet and drawer. We collected a play dataset comprising four hours of teleoperated interaction ($\sim$450{,}000 transitions) using a VR controller to guide the robot. RGB observations were recorded from both a static and a gripper-mounted camera. We evaluated the model on three representative skills: opening the drawer, closing the drawer, and pushing the cabinet slider to the right (see~\figref{fig:real_skills}). To pre-train the diffusion policies and reward classifiers, we collected 50 expert demonstrations per skill. We trained a generative world model on the offline play dataset and found that it was capable of accurate long-horizon predictions in held-out trajectories. Qualitative rollout examples are provided in the Appendix~\ref{sec:wm_rollouts}. We then used the trained world model to encode expert demonstrations into latent representations, which were used to pre-train separate diffusion policies and reward classifiers for each skill. Finally, we fine-tuned the pre-trained policies for $\sim$2 million imagination steps entirely within the latent space of the learned world model.\looseness=-1

To evaluate performance, we executed 20 rollouts per skill using fixed initial scene configurations and robot starting positions, both with the pre-trained and fine-tuned policies. Success rates, averaged over three random seeds, are reported in ~\figref{fig:real_adapt}. We find that although the pre-trained diffusion policies exhibit limited initial success across all three tasks, \ourmodel substantially improves their performance through offline fine-tuning within the learned world model. This demonstrates effective real-world policy adaptation without requiring any physical interaction.

\section{Conclusion}
\label{sec:conclusion}
We presented \textbf{\ourmodel}, a fully offline framework for adapting diffusion policies using learned world models. By treating the world model as a safe, data-driven simulator, \ourmodel enables reinforcement learning entirely in imagination, avoiding the cost and risk of online physical interactions. Our approach fine-tunes pre-trained diffusion policies through long-horizon rollouts in latent space, leveraging a compact and expressive representation of environment dynamics. On the CALVIN benchmark, \ourmodel achieves strong adaptation performance while requiring no additional environment interaction, demonstrating substantial gains in sample efficiency over model-free baselines. Our work provides the first empirical evidence that diffusion policies fine-tuned entirely offline within a learned world model trained on real-world unlabeled play data can transfer zero-shot to real-world robotic systems.\looseness=-1

\section{Limitations}
\label{sec:limitations}
While \ourmodel enables fully offline fine-tuning of diffusion policies and achieves strong results in both simulated and real-world settings, it has several limitations that point to promising directions for future research.

First, the framework relies on a world model trained once on offline play data, which is then frozen during fine-tuning. While this eliminates the cost and risk associated with online interactions, it also means that modeling errors or artifacts in the learned dynamics persist throughout training. These imperfections can be exploited by the policy, leading to overfitting to flaws in the model. Future work could explore hybrid approaches that combine offline training with limited online interaction, allowing the world model to be incrementally updated and corrected using real-world feedback.

Second, since fine-tuning is conducted entirely in imagination, there may be a mismatch between training performance and actual real-world behavior. Improvements observed within the world model do not always guarantee successful execution on the physical robot. Consequently, intermediate checkpoints must be evaluated on the real system to assess true performance.
%===============================================================================
%===============================================================================
% %\clearpage
% % The acknowledgments are automatically included only in the final and preprint versions of the paper.
\acknowledgments{This work was supported by the BrainWorlds initiative of the BrainLinks-BrainTools center at the University of Freiburg.}
%===============================================================================
% no \bibliographystyle is required, since the corl style is automatically used.
\bibliography{main}  % .bib
\newpage
\appendix
\setcounter{section}{0}
\setcounter{table}{0}
\setcounter{figure}{0}
\renewcommand{\thesection}{S.\arabic{section}}
\renewcommand{\thesubsection}{S.\arabic{section}.\arabic{subsection}}
\renewcommand{\thetable}{S.\arabic{table}}
\renewcommand{\thefigure}{S.\arabic{figure}}
\renewcommand{\thealgorithm}{S.\arabic{algorithm}}
{\Large{\bf Supplementary Material}}\\
% \normalsize
\section{Hyperparameters and Training Details}
\label{sec:hyperparams}
\subsection{World Model}
\begin{table}[b]
\vspace{-0.5cm}
\footnotesize
\centering
\caption{Hyperparameters used for training the world model. All values are shared across simulation and real-world experiments, except KL loss scale \( \beta \), which is 0.3 for simulation and 1.0 for real-world settings.}
\label{tab:app_wm_hparams}
\begin{tabular}{lcc}
\toprule
\textbf{Name} & \textbf{Symbol} & \textbf{Value} \\
\midrule
Batch size & $B$ & 50 \\
Sequence length & $L$ & 50 \\
Deterministic latent state dimensions & --- & 1024 \\
Discrete latent state dimensions & --- & 32 \\
Discrete latent state classes & --- & 32 \\
Latent dimensions & $k$ & 2048 \\
KL loss scale & $\beta$ & 0.3 \\
KL balancing coefficient & $\delta$ & 0.8 \\
RSSM reset probability & $\zeta$ & 0.01 \\
World model learning rate & --- & $3 \times 10^{-4}$ \\
Gradient clipping & --- & 100 \\
Adam epsilon & $\epsilon$ & $10^{-5}$ \\
Weight decay (decoupled) & --- & $5 \times 10^{-2}$ \\
\bottomrule
\end{tabular}
\end{table}
Following the design introduced in LUMOS~\cite{nematolli2024lumos}, we adopt a DreamerV2-style latent dynamics model as the backbone of our world model. While DreamerV2 was originally proposed for Atari game environments~\cite{dreamerv2}, our setting focuses on robotic manipulation using raw teleoperated play data. To accommodate this domain shift, we integrate two separate visual encoders for the static and wrist-mounted gripper cameras. Their encoded features are concatenated and fused via a fully-connected layer before being passed to the recurrent state-space model (RSSM). This fusion allows the model to jointly reason over both ego-centric and third-person viewpoints during prediction and imagination. Our world model is trained by minimizing the negative variational Evidence Lower Bound (ELBO):
\begin{equation}
\label{eq:app_elbo}
\min_{\phi}\; \mathbb{E}_{q_\phi}\left[
\sum_{t=1}^T -\log p_\phi(x_t \mid s_t) + \beta \, \text{KL}\left(q_\phi(z_t \mid h_t, x_t)\, \| \, p_\phi(z_t \mid h_t)\right)
\right],
\end{equation}
where \( s_t = (h_t, z_t) \), and \( \beta \) controls the strength of KL regularization. To stabilize learning, we apply KL balancing to modulate gradient flow between the prior and posterior distributions, following the formulation from~\citet{dreamerv2}:
\begin{equation}
\text{KL}(q \parallel p) = \delta \underbrace{\text{KL}(q \parallel \text{sg}(p))}_{\text{posterior regularizer}} + (1-\delta)\underbrace{\text{KL}(\text{sg}(q) \parallel p)}_{\text{prior regularizer}},
\end{equation}
where \( \text{sg}(\cdot) \) denotes the stop-gradient operator. We found KL balancing to be crucial for improving the sharpness and consistency of imagined rollouts, as it accelerates the prior’s convergence toward the richer posterior distribution.

The stochastic latent code \( z_t \) is modeled using a discrete representation composed of 32 categorical variables with 32 possible classes each. This leads to a sparse 1024-dimensional one-hot vector, which we concatenate with the deterministic hidden state \( h_t \) of size 1024, yielding a total latent dimensionality of \( k = 2048 \). We train all components of the world model jointly using sequences of 50 steps sampled from diverse points in long-horizon play episodes. Due to the scarcity of resets in such data, we reset the recurrent state of the RSSM with a small probability \( \zeta \) to encourage robustness to initialization and better exploitation of temporal context. All hyperparameters are kept identical across simulation and real-world experiments, except for the KL loss scale \( \beta \), which is set to 0.3 in simulation and 1.0 in real-world training.
To maximize coverage of different scene transitions, we sample training subsequences by selecting random start indices within each episode, ensuring the sampled subsequence remains within episode bounds. This configuration is used consistently across both simulated and real-world settings unless otherwise noted (See \tabref{tab:app_wm_hparams}).

\subsection{Diffusion Policy}
We adopt a denoising diffusion probabilistic model (DDPM)~\cite{ddpm} to parameterize our base policy. The diffusion policy is trained to imitate expert trajectories using features produced by our frozen world model encoder. Specifically, we featurize each raw observation with the world model to obtain 2048-dimensional latent vectors, which serve as the input to the policy $\pi_\theta(\cdot \mid z_t)$. This featurization ensures compatibility between the policy's training and inference regimes, as the fine-tuned policy will later be conditioned on imagined future latent states.

For each skill, we use \( N = 50 \) expert demonstration trajectories, randomly selected from task-annotated episodes in the CALVIN simulation~\cite{mees2022calvin} and manually collected in the real-world environment. The diffusion model is trained with $K = 20$ denoising steps, and follows a chunked prediction strategy: given an observation horizon of 1 step, it predicts a sequence of $T_p = 4$ future actions, of which the first $T_a = 4$ are executed in the environment. The policy is optimized using a behavior cloning objective over the full denoising trajectory:
\begin{equation}
\label{eq:app_dp_bc}
\mathcal{L}_\text{BC}(\theta) = \mathbb{E}_{\mathcal{D}_\text{exp}} \left[
\sum_{t=1}^{T} \sum_{k=1}^{K} 
- \log \pi_\theta(a_t^{k-1} \mid z_t, a_t^k)
\right],
\end{equation}
where \( \pi_\theta \) predicts denoised actions conditioned on the current latent state \( z_t \) and noisy action \( a_t^k \).

The policy model is a multi-layer perceptron (MLP) with three hidden layers of size 512, and we apply exponential moving average (EMA) to the policy weights during training, starting from epoch 20, to enhance stability~\cite{morales2024exponential}. All policies are trained for 5000 epochs using the Adam optimizer. We use an initial learning rate of \( 1 \times 10^{-4} \), decayed to \( 1 \times 10^{-5} \) using a cosine schedule. We apply a weight decay of \( 1 \times 10^{-6} \) and use a batch size of 256. These hyperparameters are kept identical across all CALVIN tasks and our real-world skill evaluations (See \tabref{tab:app_dp_hparams}).
\begin{table}[b]
\vspace{-0.5cm}
\footnotesize
\centering
\caption{Training and model hyperparameters for diffusion policy across all CALVIN and real-world tasks.}
\label{tab:app_dp_hparams}
\begin{tabular}{lcc}
\toprule
\textbf{Parameter} & \textbf{Symbol} & \textbf{Value} \\
\midrule
\multicolumn{3}{l}{\textit{Common Training Parameters (All Skills)}} \\
Observation Horizon & --- & 1 \\
Number of Demonstrations & $N$ & 50 \\
Planning Horizon & $T_p$ & 4 \\
Action Horizon & $T_a$ & 4 \\
Training Epochs & --- & 5000 \\
Diffusion Denoising Steps & $K$ & 20 \\
Initial Learning Rate & --- & $1 \times 10^{-4}$ \\
Final Learning Rate & --- & $1 \times 10^{-5}$ \\
Weight Decay & --- & $1 \times 10^{-6}$ \\
MLP Dimensions & --- & {[}512, 512, 512{]} \\
EMA Decay & --- & 0.995 \\
EMA Start Epoch & --- & 20 \\
EMA Update Frequency & --- & 10 \\
Batch Size & --- & 256 \\
\midrule
\multicolumn{3}{l}{\textit{Observation Dimensions}} \\
\ourmodel & --- & 2048 \\
DPPO (Vision WM Encoder) & --- & 2048 \\
DPPO (Vision) & --- & $64\times64\times6$ \\
DPPO (State) & --- & 51 \\
\bottomrule
\end{tabular}
\end{table}

When evaluating the DPPO baseline in the CALVIN simulation environment, we also include a variant that has access to ground-truth state information, which has an observation dimensionality of 51. For the vision-based variant, the input consists of RGB images from both the static and gripper cameras, stacked along the channel dimension, resulting in an input shape of \(64 \times 64 \times 6\).\looseness=-1

\subsection{Latent Reward Estimator}
\label{sec:reward_clf}
To learn a task-aligned reward signal, we train a latent reward classifier \( C_\psi \) using expert demonstration data \( \mathcal{D}_\text{exp} \). Each observation \( x_t \) is encoded into a latent state \( z_t \) via the frozen world model encoder. The classifier comprises two components: a two-layer MLP \( f_\psi \) that maps latents to an embedding space, and a subsequent two-layer MLP \( g_\psi \) that predicts success or failure based on the embedding.

We jointly optimize the model using a combination of contrastive and classification losses. For the contrastive component, we employ the NT-Xent loss~\cite{sohn2016improved}, which encourages embeddings of positive pairs to be closer than those of negative pairs. Given a batch of \( N \) samples, the NT-Xent loss for a positive pair \( (i, j) \) is defined as:
\begin{equation}
\label{eq:nt_xent}
\mathcal{L}_{\text{NT-Xent}} = -\log \frac{\exp(\text{sim}(f_\psi(z_i), f_\psi(z_j))/\tau)}{\sum_{k=1}^{2N} \mathbb{1}_{[k \ne i]} \exp(\text{sim}(f_\psi(z_i), f_\psi(z_k))/\tau)},
\end{equation}
where \( \text{sim}(\cdot, \cdot) \) denotes the cosine similarity, \( \tau \) is a temperature parameter, and \( \mathbb{1}_{[k \ne i]} \) is an indicator function excluding the anchor sample from the denominator.

In parallel, the classification MLP \( g_\psi \) operates on the embeddings to predict success labels, trained using standard cross-entropy loss. The overall training objective combines both terms:
\begin{equation}
\label{eq:app_reward_loss}
\mathcal{L}_{\text{reward}} = \mathcal{L}_{\text{NT-Xent}} + \mathcal{L}_{\text{CE}}.
\end{equation}

The resulting reward function is defined as \( R_\psi(z_t, a_t):= \text{softmax}(g_\psi(f_\psi(z_t))) \), which outputs the predicted probability of success given a latent observation. Both MLPs use ReLU activations, and the model is trained with the Adam optimizer for 100 epochs.  See Table~\ref{tab:app_reward_hparams} for the full set of hyperparameters.

In practice, leveraging the world model’s structured latent states allows the reward classifier to achieve high accuracy with as few as 50 demonstrations per task: we treat successful states as positives and randomly sample 15\% of the remaining frames as negatives, yielding an average of 0.89 precision and 0.98 recall across eight CALVIN skills. A vision-based ResNet-18 trained on the same data matches recall but achieves only 0.41 precision, underscoring that robustness primarily stems from the temporally structured latent space of the world model (see Table~\ref{tab:reward_cls_perf} for details). \looseness=-1

\begin{table}[h]
\centering
\setlength{\tabcolsep}{4.5pt}
\footnotesize
\begin{minipage}[t]{0.38\linewidth}
\caption{Hyperparameters used for training the latent reward classifier.}
\label{tab:app_reward_hparams}
\centering
\begin{tabular}{lc}
\toprule
\textbf{Parameter} & \textbf{Value} \\
\midrule
Embedding MLP & {[}512, 512{]} \\
Classification MLP & {[}512, 512{]} \\
Activation & ReLU \\
Output & Softmax \\
Epochs & 100 \\
Batch Size & 32 \\
Learning Rate & $1\times10^{-6}$ \\
Temperature & 0.5 \\
Loss & Contrastive+CE \\
Positives & Success frames \\
Negatives & 15\% other frames \\
\bottomrule
\end{tabular}
\end{minipage}
\hspace{0.05\linewidth}% <-- add fixed horizontal space here
\begin{minipage}[t]{0.55\linewidth} % slightly narrower
\caption{Validation precision and recall for our latent-based and vision-based reward classifiers.}
\label{tab:reward_cls_perf}
\centering
{\setlength{\tabcolsep}{3.5pt} % reduce horizontal padding
\begin{tabular}{c@{}|cc|cc@{}}
\toprule
\textbf{Task} & \multicolumn{2}{c|}{\textbf{Latents}} & \multicolumn{2}{c}{\textbf{Vision}} \\
              & \textbf{Prec.} & \textbf{Rec.} & \textbf{Prec.} & \textbf{Rec.} \\ \midrule
\odrawer      & 0.92 & 0.99 & 0.41 & 0.99 \\
\cdrawer      & 0.89 & 0.99 & 0.52 & 0.99 \\
\lslider      & 0.87 & 0.96 & 0.33 & 0.97 \\
\rslider      & 0.83 & 0.98 & 0.41 & 0.99 \\
\lighton      & 0.89 & 0.96 & 0.45 & 0.99 \\
\lightoff     & 0.88 & 0.99 & 0.36 & 1.00 \\
\ledon        & 0.94 & 1.00 & 0.36 & 0.92 \\
\ledoff       & 0.88 & 0.99 & 0.41 & 0.99 \\ \midrule
\textbf{Avg.} & \textbf{0.89} & \textbf{0.98} & 0.41 & \textbf{0.98} \\ \bottomrule
\end{tabular}
}
\end{minipage}
\end{table}

\subsection{Fine-tuning with \ourmodel}
\begin{algorithm}[t]
\caption{\texttt{\ourmodel}: Diffusion Policy Adaptation with World Models}
\label{alg:diwa}
\begin{algorithmic}[1]
\State Train world model $M_\phi$ on play data $\mathcal{D}_\text{play}$ using the ELBO objective (Eq.~\eqref{eq:app_elbo}), then freeze $M_\phi$.
\State Encode expert demonstrations into latents $z_t \sim q_\phi(z_t \mid h_t, x_t)$ using the frozen world model.
\State Pre-train diffusion policy $\pi_\theta$ on latent expert demonstrations via behavior cloning (Eq.~\eqref{eq:app_dp_bc}); freeze copy as $\pi_{\theta_\text{pre}}$.
\State Train reward classifier $C_\psi$ on latent expert demonstrations via reward loss (Eq.~\eqref{eq:app_reward_loss}).
\State Initialize value function $V_\nu$.
\For{iteration = 1, 2, \dots}
    \State Initialize imagined rollout buffer $\mathcal{D}_\text{itr}$.
    \State Set $\pi_{\theta_\text{old}} = \pi_\theta$.
    \For{imagination episode = 1, 2, \dots, $N$ in parallel}
        \State Sample initial observation $x_0$ and encode to latent $z_0$.
        \State Initialize state $\bar{s}_{\bar{t}(0,K)} = (z_0, \bar{a}_0^K)$ in $\mathcal{M}_{\text{DD}}$.
        \For{imagined step $t = 0, \dots, T-1$, denoising step $k = K, \dots, 1$}
            \State Sample intermediate action $\bar{a}_t^{k-1} \sim \bar{\pi}_{\theta_\text{old}}(\cdot \mid z_t, \bar{a}_t^k)$
            \If{$k = 1$}
                \State Run final action $\bar{a}_t^0$ in the world model $M_\phi$
                \State Update recurrent state: $h_{t+1} = f_\phi(h_t, \bar{a}_t^0)$
                \State Sample next latent state: $z_{t+1} \sim p_\phi(z_{t+1} \mid h_{t+1})$
                \State Predict reward: $\bar{R}_{\bar{t}(t,1)} = R_\psi(z_t, \bar{a}_t^0)$
                \State Sample new noisy action: $\bar{a}_{t+1}^{K} \sim \mathcal{N}(0, I)$
                \State Set next state: $\bar{s}_{\bar{t}(t+1,K)} = (z_{t+1}, \bar{a}_{t+1}^K)$
            \Else
                \State Set reward: $\bar{R}_{\bar{t}(t,k)} = 0$
                \State Set next state: $\bar{s}_{\bar{t}(t,k-1)} = (z_t, \bar{a}_t^{k-1})$

            \EndIf
            \State Add $(k, \bar s_{\bar t(t, k)}, \bar a_{\bar t(t, k)}, \bar R_{\bar t(t, k)})$ to $\mathcal D_\text{itr}$.
        \EndFor

    \EndFor
    \State Compute advantage estimates $A^{\pi_{\theta_\text{old}}}(\bar{s}_{\bar{t}(t,1)}, \bar{a}_{\bar{t}(t,1)})$ using GAE (Eq.~\eqref{eq:app_gae})
    \For{update = 1, \dots, $\text{num\_updates}$}
        \For{minibatch = 1, \dots, $B$}
            \State Sample $(k, \bar s_{\bar t(t, k)}, \bar a_{\bar t(t, k)}, \bar R_{\bar t(t, k)})$ and $A^{ \pi_{\theta_\text{old}}}(s_{\bar t(t, k)}, a_{\bar t(t, k)})$ from $\mathcal D_\text{itr}$.
            \State Compute denoising-discounted advantage $\hat A_{\bar t(t, k)} = \gamma_\text{denoise}^k A^{ \pi_{\theta_\text{old}}}(s_{\bar t(t, 0)}, a_{\bar t(t, 0)})$. 
            \State Update $\pi_\theta$ using regularized PPO loss (Eq.~\eqref{eq:app_ppo_bc}).
            \State Update $V_\nu$ using value loss (Eq.~\eqref{eq:app_value}).
        \EndFor
    \EndFor
\EndFor
\State \Return fine-tuned policy $\pi_\theta$.
\end{algorithmic}
\end{algorithm}
The full pseudocode for \ourmodel is shown in Algorithm~\ref{alg:diwa}. \ourmodel fine-tunes a pre-trained diffusion policy $\pi_\theta$ using imagined rollouts from a learned world model $M_\phi$ and reward classifier $C_\psi$, forming trajectories in the Dream Diffusion MDP $\mathcal{M}_\text{DD}$. At each iteration, imagined transitions are stored in a buffer $\mathcal{D}_\text{itr}$, advantages are estimated using Generalized Advantage Estimation (GAE)~\citep{schulman2015high}, and PPO-style updates~\citep{schulman2017proximal} are applied to the policy and value function. GAE is computed at the final denoising step ($k = 1$) for each world model timestep:
\begin{equation} 
\label{eq:app_gae}
    \hat{A}_{\bar t(t,1)}^\lambda = \sum_{l=0}^{\infty} (\gamma_\text{WM} \lambda)^l \bar \delta_{\bar t(t+l,1)}, \quad \text{where } \bar \delta_{\bar t(t,1)} = \bar R_{\bar t(t,1)} + \gamma_\text{WM} V_\nu(\bar s_{\bar t(t+1,1)}) - V_\nu(\bar s_{\bar t(t,1)}).
\end{equation}
To propagate this signal to earlier denoising steps, we apply a denoising discount to obtain step-specific advantages as $\hat{A}_{\bar{t}(t,k)} = \gamma_\text{denoise}^k \hat{A}_{\bar t(t,1)}$. The policy is fine-tuned using a behavior-regularized PPO objective that augments the clipped PPO loss with a behavior cloning (BC) regularization term. This regularization encourages proximity to the pre-trained diffusion policy $\pi_{\theta_\text{pre}}$, mitigating overfitting to model errors during imagination~\citep{modelexploit, torne2024reconciling}. The full objective is:
\begin{equation} 
\label{eq:app_ppo_bc}
\mathcal{L}_\theta = \mathcal{L}_\text{PPO} - \alpha_\text{BC} \ \mathbb{E}^{\bar{\pi}_{\theta_\text{old}}} \left[\sum_{k=1}^{K} \log \pi_{\theta_\text{pre}}(\bar{a}_t^{k-1} \mid z_t, \bar{a}_t^k)\right],
\end{equation}
where $\alpha_\text{BC}$ controls the regularization strength and $\pi_{\theta_\text{pre}}$ remains frozen during fine-tuning. To restrict updates to the last \( K' \) denoising steps, we subsample \( \mathcal{D}_\text{itr} \) to include only entries with \( k \leq K' \), keeping the base policy \( \pi_{\theta_\text{pre}} \) frozen for the initial \( K - K' \) steps.
The value function $V_\nu$ is trained to regress the future discounted sum of latent rewards:
\begin{equation}
\label{eq:app_value}
\mathcal{L}_\nu = \mathbb{E}_{\mathcal{D}_\text{itr}} \left[ \left( \sum_{l=0}^{T-t} \gamma_\text{WM}^l \bar R_{\bar t(t+l,1)} - V_\nu(z_t) \right)^2 \right],
\end{equation}
where $V_\nu$ takes as input only the latent state $z_t$ from the $\mathcal{M}_\text{DD}$. \tabref{tab:app_finetune_hyperparams} lists the fine-tuning hyperparameters shared across all skills and experiments for both \ourmodel and the baseline methods. We set the behavior cloning regularization coefficient \(\alpha_{\text{BC}} = 0.05\) for all tasks by default, except for \odrawer{}, \cdrawer{}, and \ledon{}, where we observed better performance with values of \(0.10\), \(0.025\), and \(0.025\), respectively.
\looseness=-1
\begin{table}[t]
\footnotesize
\centering
\vspace{-0.5cm}
\caption{Fine-tuning hyperparameters shared across all skills for \ourmodel and baseline methods.}
\label{tab:app_finetune_hyperparams}
\begin{tabular}{lll}
\toprule
\textbf{Parameter} & \textbf{Symbol} & \textbf{Value} \\
\midrule
Planning Horizon (Environment) & $T_p$ & 4 \\
Planning Horizon (Actor) & $T_a$ & 4 \\
Denoising Steps & $K$ & 20 \\
Fine-tuned Denoising Steps& $K'$ & 10 \\
Actor Learning Rate & --- & $1 \times 10^{-5}$ \\
Critic Learning Rate & --- & $1 \times 10^{-3}$ \\
Actor MLP Dimensions & --- & {[}512, 512, 512{]} \\
Critic MLP Dimensions & --- & {[}256, 256, 256{]} \\
Discount Factor (Env /World Model) & $\gamma_{\text{ENV}}$ / $\gamma_{\text{WM}}$ & 0.999 \\
Discount Factor (Diffusion Policy) & $\gamma_{\text{DP}}$ & 0.99 \\
GAE Smoothing Parameter & $\lambda$ & 0.95 \\
Behavior Cloning Coefficient (default) & $\alpha_{\text{BC}}$ & 0.05 \\
Batch Size & --- & 7500 \\
\bottomrule
\vspace{-0.5cm}
\end{tabular}
\end{table}
\section{Experimental Setup Details}
\subsection{7-DoF Action Framework}
All experiments, both in simulation and in the real world, use a 7-dimensional action space defined as:\looseness=-1
\[
[\delta x, \delta y, \delta z, \delta \phi, \delta \theta, \delta \psi, \textit{gripperAction}]
\]
The first six dimensions control the end-effector, with $(\delta x, \delta y, \delta z)$ specifying position changes and $(\delta \phi, \delta \theta, \delta \psi)$ specifying orientation changes via Euler angles. Each takes continuous values in the range $[-1, 1]$. The final dimension, \textit{gripperAction}, controls the gripper state. Although the environment expects discrete inputs ($1.0$ to close, $-1.0$ to open), \ourmodel{} outputs a continuous value in $[-1.0, 1.0]$, which is thresholded before execution: values greater than or equal to 0 trigger opening, and values less than 0 trigger closing.
\subsection{Real-World Data Collection}
\begin{figure*}[t]
    \centering
    \subfloat[Real-World Setup]{
        \includegraphics[width=7cm]{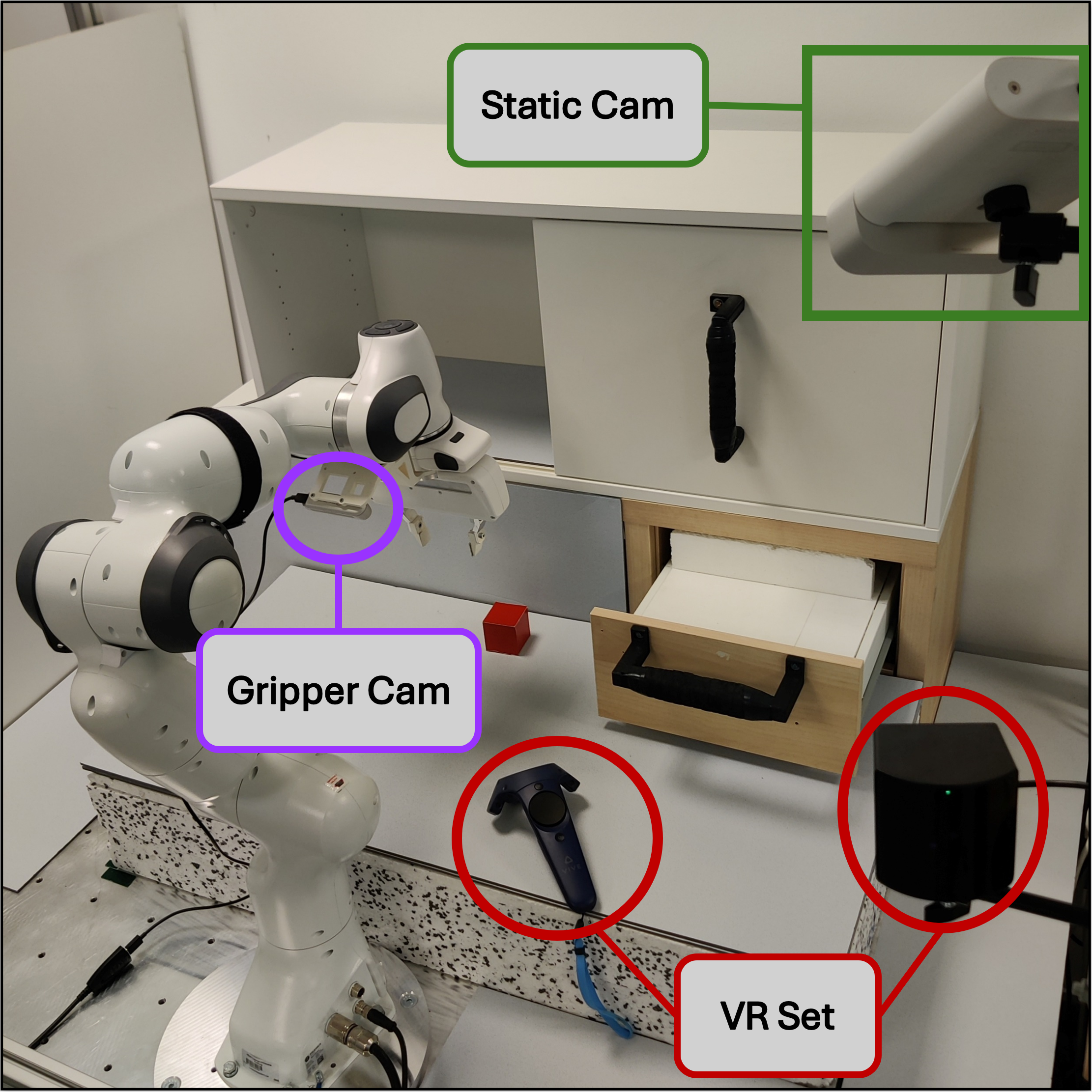}\label{fig:app_real_setup}}
    \subfloat[Real-World Observations]{
        \includegraphics[width=7cm]{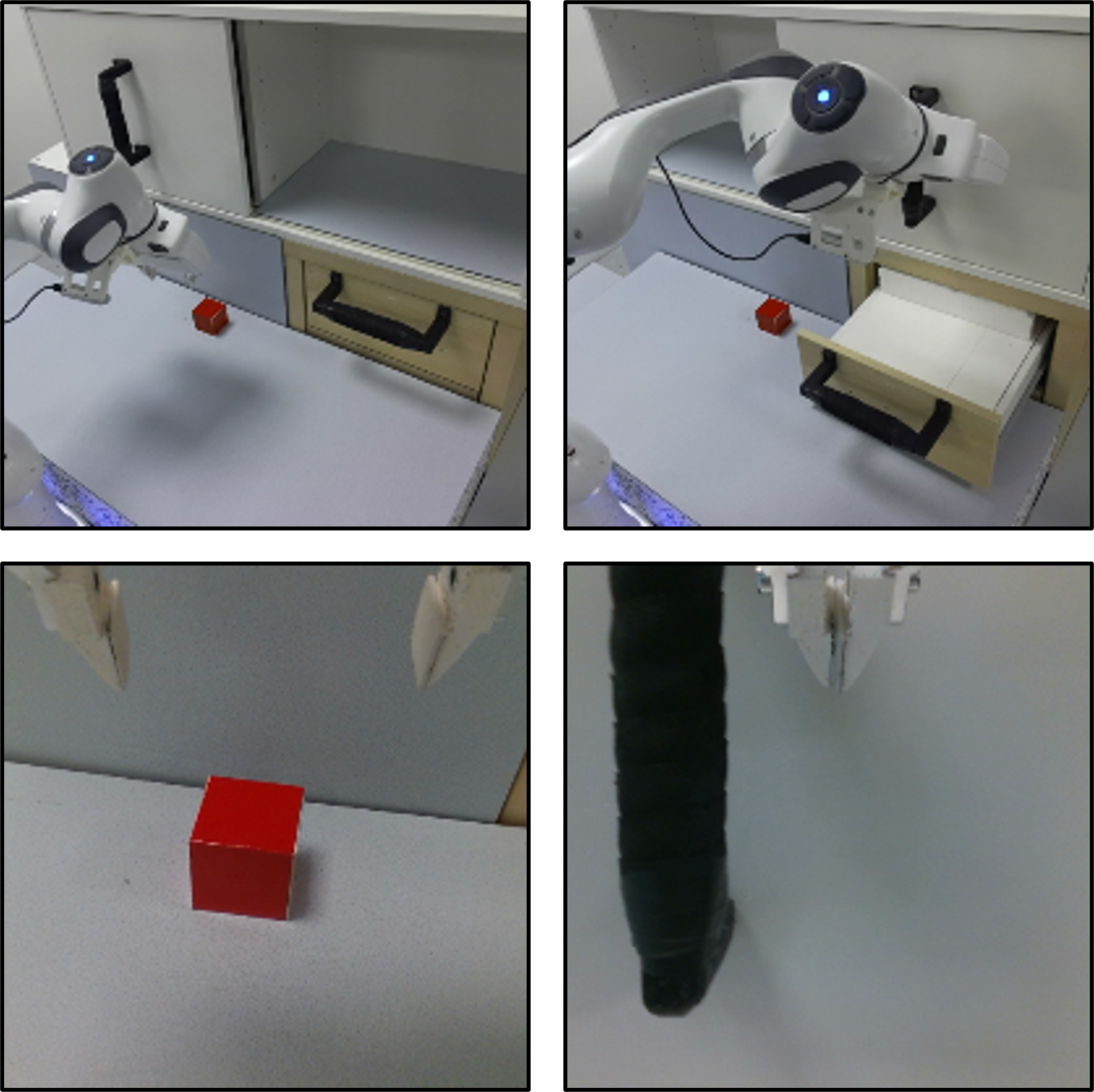}\label{fig:app_real_obs}}
    \caption{(a) Real-world setup showing the Franka Panda robot, VR teleoperation interface (HTC VIVE controller and tracking system), and camera placements (static Kinect and wrist-mounted Realsense). (b) Example observations from the static and gripper-mounted RGB cameras used during data collection.}
    \label{fig:real-setup-obs}
\end{figure*}

We collected four hours of real-world teleoperation data using a Franka Emika Panda robot controlled via an HTC VIVE Pro headset in a 3D tabletop setting (see Figure~\ref{fig:app_real_setup}). The tabletop environment included a cabinet with a drawer and a manipulable red cube to support diverse interaction scenarios. During teleoperation, we recorded robot sensor data, including proprioceptive signals (joint states and end-effector pose), as well as multimodal visual observations. RGB images of the full scene were captured at a resolution of \(200 \times 200\) using an Azure Kinect camera, while close-up RGB views of the manipulated objects were obtained from a wrist-mounted Realsense D415 camera (Figure~\ref{fig:app_real_obs}). We also logged the absolute control commands sent to the robot. For model training, we computed relative actions as differences between consecutive absolute commands. To reduce redundancy caused by low inter-frame variation, the original 30 Hz recording rate was downsampled by a factor of 4 to 7.5 Hz.\looseness=-1
\section{Data Preprocessing}
In both simulation and real-world experiments, we use visual observations from two sources: a static camera and a wrist-mounted gripper camera. All images are first resized to a resolution of \(64 \times 64\) pixels. We then convert the image tensors from integer values in \([0, 255]\) to floating-point values in \([0.0, 1.0]\), and subsequently normalize them. These transformations are applied to both static and gripper observations. In addition to visual observations, we preprocess the robot state, which includes the end-effector's position and orientation. Since the orientation is originally represented in Euler angles, we convert it to a continuous 6D rotation representation~\cite{zhou2019continuity} to avoid discontinuities and singularities associated with Euler angles.

\section{Additional Experiments}
\begin{figure}[t]
    % \vspace{-0.5cm}
    \centering
    \includegraphics[width=1.0\linewidth]{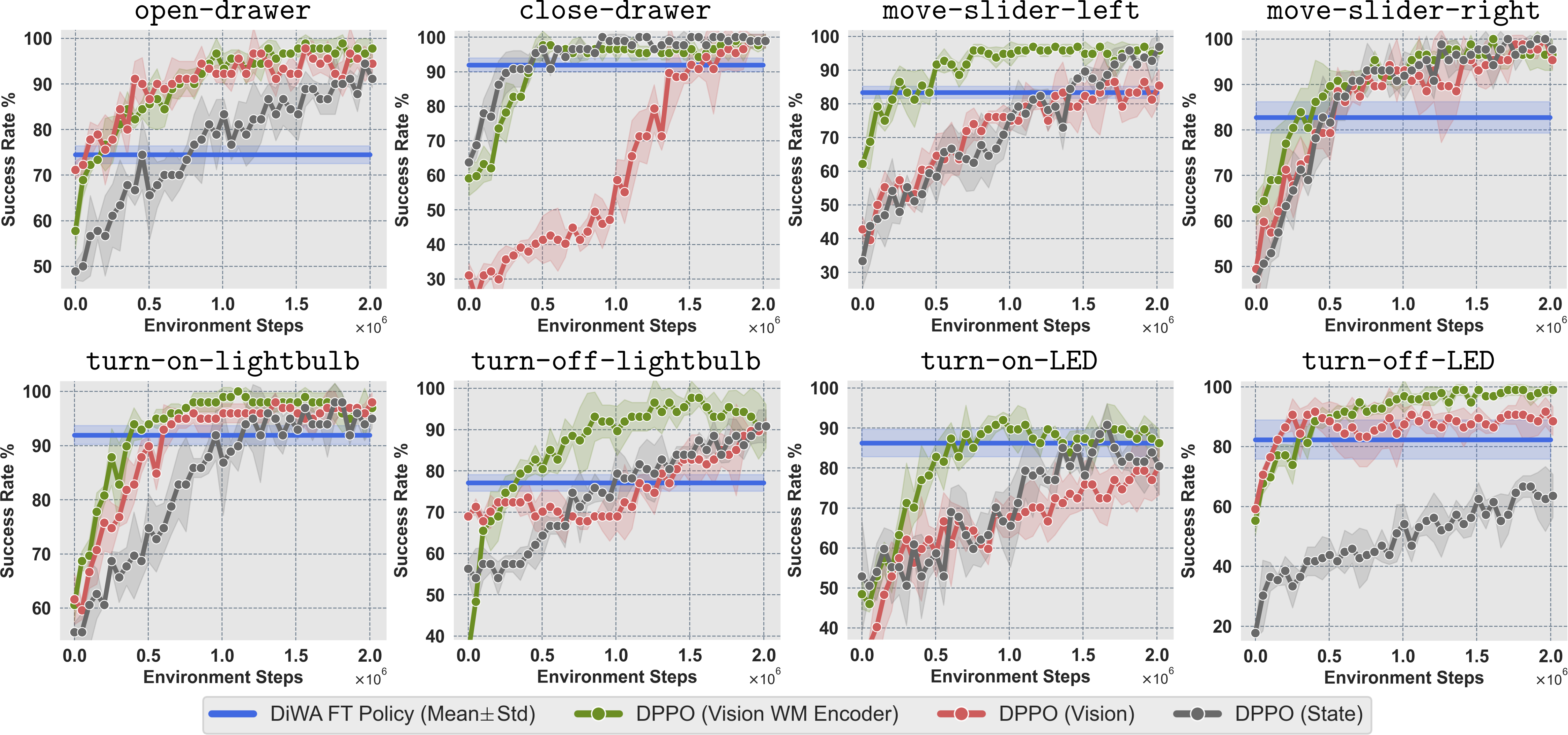}
    \caption{Comparison of \ourmodel with three DPPO variants using different input modalities. \ourmodel{} (blue) fine tunes policies entirely offline using a learned world model, requiring no physical interaction during adaptation. In contrast, DPPO (gray, red, green) performs online reinforcement learning with access to environment rewards and dynamics. The DPPO variant using latents from the world model encoder (green) achieves the highest performance among the three, but all require hundreds of thousands of real world interactions per skill.\looseness=-1}
    \label{fig:dppo-online}
    % \vspace{-0.3cm}
\end{figure}
\subsection{Comparing DPPO Input Modalities}
\label{sec:dppo_variants}

Figure~\ref{fig:dppo-online} compares three DPPO configurations against our offline method. DPPO (State) (gray) uses raw simulator state as input, DPPO (Vision) (red) operates directly on pixel observations using a Vision Transformer (ViT) based encoder~\cite{hu2023imitation}, and DPPO (Vision WM Encoder) (green) uses visual inputs processed through the same frozen encoder employed in our world model. Among these, the world model latent variant, where DPPO operates on representations produced by our recurrent state space model, often achieves the highest performance, surpassing both raw vision and state-based inputs. These latents combine a history-aware deterministic hidden state with a stochastic component that captures residual uncertainty, providing a compact and dynamics-aligned representation. In contrast to all online variants, \ourmodel{} (blue) fine-tunes the policy entirely offline using imagined rollouts in the learned latent space. Its performance is shown as a horizontal band, as no physical interaction is required during fine-tuning. While DPPO can eventually match or exceed our results by leveraging ground truth dynamics and rewards, it requires \textbf{hundreds of thousands} of real-world interactions per skill. These interactions are costly, time-consuming, and can pose safety risks. In comparison, \ourmodel{} achieves competitive results using only a few hours of play data, offering a safer and more sample-efficient approach to real-world skill adaptation.
% \vspace{-0.4cm}
\begin{figure}[b]
    \centering
    \includegraphics[width=1.0\linewidth]{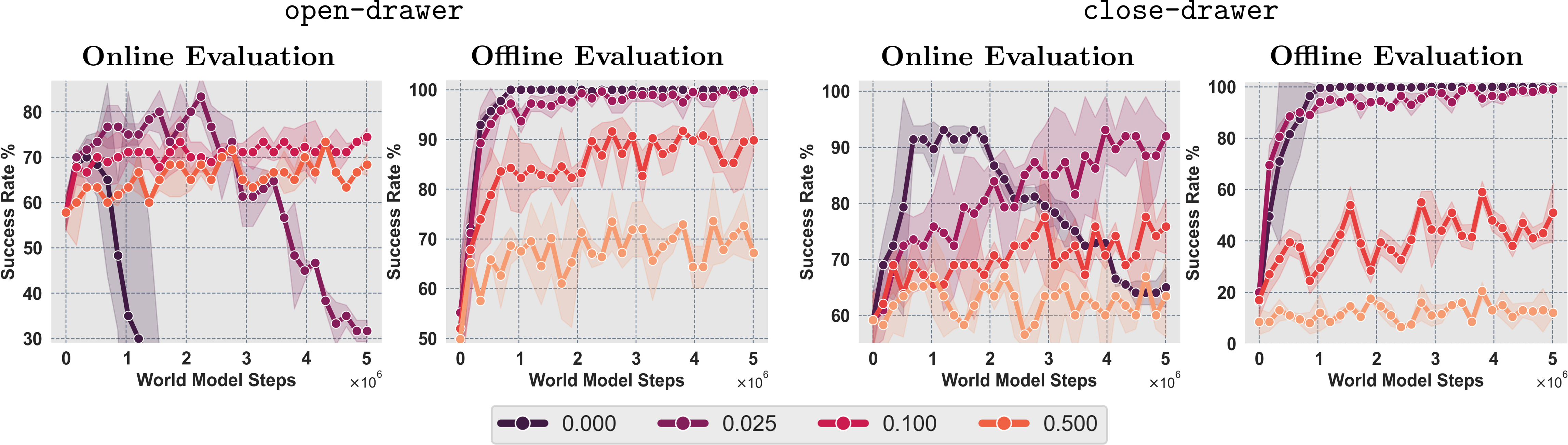}
    \caption{Ablation of behavior cloning regularization strength ($\alpha_\text{BC}$) during fine-tuning. Without regularization ($\alpha_\text{BC} = 0.0$), the agent performs well in imagination but fails in the real environment, indicating exploitation of world model inaccuracies. Excessively high values (e.g., $0.5$) prevent meaningful adaptation. Intermediate values strike a balance, yielding robust transfer.}
    \label{fig:app_bc_loss}
\end{figure}
\subsection{Impact of Behavior Cloning Regularization}
To investigate the role of behavior cloning regularization in fine-tuning, we ablate the BC loss coefficient $\alpha_\text{BC}$ in \ourmodel{} and evaluate performance across different settings. As shown in Figure~\ref{fig:app_bc_loss}, the choice of $\alpha_\text{BC}$ has a significant impact on performance.\looseness=-1

When $\alpha_\text{BC} = 0.0$, meaning no regularization is applied, the agent achieves high success rates during offline evaluation within the imagined environment. However, this performance does not transfer to the real environment, where success rates drop considerably. This discrepancy suggests that the agent overfits to inaccuracies in the world model by exploiting artifacts that yield high imagined rewards but do not correspond to meaningful success in reality~\cite{modelexploit}. On the other hand, setting $\alpha_\text{BC}$ too high, such as 0.5, leads to minimal improvement over the pre-trained policy. In this case, strong regularization prevents the policy from effectively adapting to new task-specific feedback, resulting in stagnated learning. Moderate values of $\alpha_\text{BC}$ provide a better trade-off, enabling the policy to adapt while still maintaining alignment with the pre-trained behavior. These results emphasize the importance of tuning BC regularization to balance adaptation and stability when fine-tuning policies with learned world models.\looseness=-1

This issue is further compounded by the fact that the world model is trained once on offline play data and remains fixed during fine-tuning. While this avoids the cost and risk of real-world interactions, any modeling errors or artifacts in the learned dynamics persist and may be exploited by the policy. Future work could explore hybrid approaches that incorporate limited online interaction, allowing the world model to be gradually refined with real-world feedback and reducing the impact of such artifacts.\looseness=-1

\subsection{Fine-tuning a Unimodal Gaussian Policy}
While the primary focus of this work is on fine-tuning diffusion policies, which involve long denoising sequences that make reward propagation particularly difficult, our method is not limited to this specific policy class. To demonstrate the generality of our formulation, we replace the diffusion policy in \ourmodel{} with a unimodal Gaussian policy parameterized by a mean and a diagonal covariance. Unlike diffusion policies, this architecture yields a much shorter Markov chain, allowing reward signals and policy gradients from PPO to propagate more directly. As shown in \tabref{tab:app_gaussian_policy}, our fine-tuning procedure leads to consistent improvements across all tasks. This supports the claim that the underlying world model MDP, including the reward estimation mechanism, is independent of the policy architecture.\looseness=-1
\setlength{\tabcolsep}{4.5pt}
\begin{table}[h]
\footnotesize
\centering
\caption{Offline fine-tuning improves a unimodal Gaussian policy across all tasks. Success rates increase substantially without any additional real-world interaction.}
\label{tab:app_gaussian_policy}
\begin{tabular}{c|cc}
\toprule
\textbf{Task} & \multicolumn{2}{c}{\textbf{Gaussian Policy}} \\ 
\cmidrule(lr){2-3}
             & \textbf{Pre-Trained} & \textbf{Offline Fine-Tuned} \\ \midrule
\odrawer     & $50.00 \pm 0.09$     & $\textbf{71.67} \pm \textbf{2.36}$ \\
\cdrawer     & $55.17 \pm 0.18$     & $\textbf{98.28} \pm \textbf{2.44}$ \\
\lslider     & $54.86 \pm 4.70$     & $\textbf{82.64} \pm \textbf{1.59}$ \\
\rslider     & $55.52 \pm 0.78$     & $\textbf{87.93} \pm \textbf{7.31}$ \\
\lighton     & $54.55 \pm 3.03$     & $\textbf{95.96} \pm \textbf{1.75}$ \\
\lightoff    & $62.07 \pm 4.88$     & $\textbf{77.59} \pm \textbf{2.44}$ \\
\ledon       & $44.83 \pm 0.50$     & $\textbf{77.59} \pm \textbf{7.31}$ \\
\ledoff      & $40.94 \pm 3.98$     & $\textbf{79.69} \pm \textbf{2.21}$ \\ 
\midrule
\multicolumn{2}{c}{\textbf{Total Physical Interactions:}} & \textbf{0} \\
\bottomrule
\end{tabular}
\vspace{-0.5cm}
\end{table}

% LIBERO
\subsection{Results on LIBERO-90}
\label{sec:libero}
\begin{table}[b]
\footnotesize
\centering
\caption{\ourmodel improves performance on four LIBERO-90 kitchen tasks, with results averaged over three random seeds.}
\label{tab:sim_exp_libero}
\begin{tabular}{cc|c}
\toprule
\multicolumn{1}{c|}{\multirow{3}{*}{\textbf{Task}}} & \multirow{1}{*}{\textbf{Base}} & \multicolumn{1}{c}{\textbf{DiWA (Ours)}} \\ \cmidrule{2-3}
\multicolumn{1}{c|}{} & \textbf{Diffusion Policy} & \textbf{Offline Fine-Tuning} \\  \cmidrule{2-3}
\multicolumn{1}{c|}{} & \multicolumn{2}{c}{\textbf{Success Rate}} \\ \midrule
\multicolumn{1}{c|}{\liberoodrawer}   & $40.67 \pm 3.06$ & $\textbf{77.33} \pm \textbf{3.06}$ \\
\multicolumn{1}{c|}{\liberostove}   & $54.00 \pm 7.21$ & $\textbf{91.33} \pm \textbf{3.08}$ \\
\multicolumn{1}{c|}{\liberocbdrawer}   & $27.33 \pm 3.12$ & $\textbf{78.00} \pm \textbf{8.72}$ \\
\multicolumn{1}{c|}{\liberoctdrawer}   & $75.33 \pm 2.31$ & $\textbf{100.00} \pm \textbf{0.00}$ \\
\midrule
\multicolumn{1}{c}{\textbf{Total Physical Interactions:}} && \textbf{0}\\ \bottomrule
\end{tabular}
\vspace{-0.5cm}
\end{table}
To evaluate \ourmodel on the LIBERO simulation benchmark~\cite{liu2023libero}, we train a world model on the LIBERO-90 split, which is a curated subset of LIBERO-100 containing expert demonstrations for 90 short-horizon tasks spanning 10 kitchen scenes, 6 living rooms, and 4 study tables (see Sec. 4.2 in~\cite{liu2023libero}). Unlike CALVIN’s environment~\textit{D} with a fixed tabletop layout, LIBERO-90 provides far fewer interactions per scene, making world model learning significantly more challenging.

We focus on four kitchen skills across four scenes: \textit{open the top drawer} (\liberoodrawer, scene 1), \textit{turn on the stove} (\liberostove, scene 3), \textit{close the bottom drawer} (\liberocbdrawer, scene 4), and \textit{close the top drawer} (\liberoctdrawer, scene 5). Table~\ref{tab:sim_exp_libero} reports the average success rates over three seeds. Despite the sparse data and suboptimal world model training conditions, \ourmodel successfully fine-tunes all four skills entirely offline, without additional physical interactions. We observed that different tasks required varying fine-tuning horizons to achieve stable improvement without model exploitation: \liberoodrawer\ and \liberoctdrawer\ were fine-tuned for 3M steps, \liberostove\ for 2M, and \liberocbdrawer\ for 1M.

\subsection{Offline RL Limitations in Our Setting}

Standard offline RL methods are fundamentally ill-suited to our setting. They assume (i) fully labeled, task-specific reward signals and (ii) sufficient coverage of high-value state-action regions in a fixed dataset. In contrast, \ourmodel operates on task-agnostic play data with sparse expert demonstrations and estimated rewards, violating both assumptions.

To include an offline RL baseline, we experimented with CQL~\citep{kumar2020conservative} on play data labeled using a ResNet-18 reward classifier trained from expert demonstrations. This heuristic produced a noisy reward signal (high recall  of 0.98 but low precision of 0.41; see Table~\ref{tab:reward_cls_perf}), and we segmented the continuous play streams into pseudo-episodes to enable critic training. Despite these adjustments, all offline RL runs diverged rapidly due to: (i) reward mislabeling, which caused the Q-function to propagate spurious positive values; (ii) sparse coverage of successful behaviors, preventing the critic from generalizing; and (iii) value extrapolation errors, leading to policy collapse. These results highlight the inherent incompatibility of critic-based offline RL with our setting. Consequently, we view a direct comparison as unfair to offline RL methods, whereas DiWA’s on-policy imagination with latent rewards naturally avoids these failure modes and consistently improves skills without any additional interaction.
% These results confirm that critic-based offline RL is inherently unstable in our regime, whereas DiWA’s on-policy imagination with latent rewards avoids these failure modes and reliably improves skills without any additional interaction.

\subsection{World Model Rollouts in the Real World}
\label{sec:wm_rollouts}
\begin{figure}[]
    \centering
    \includegraphics[width=1.0\linewidth]{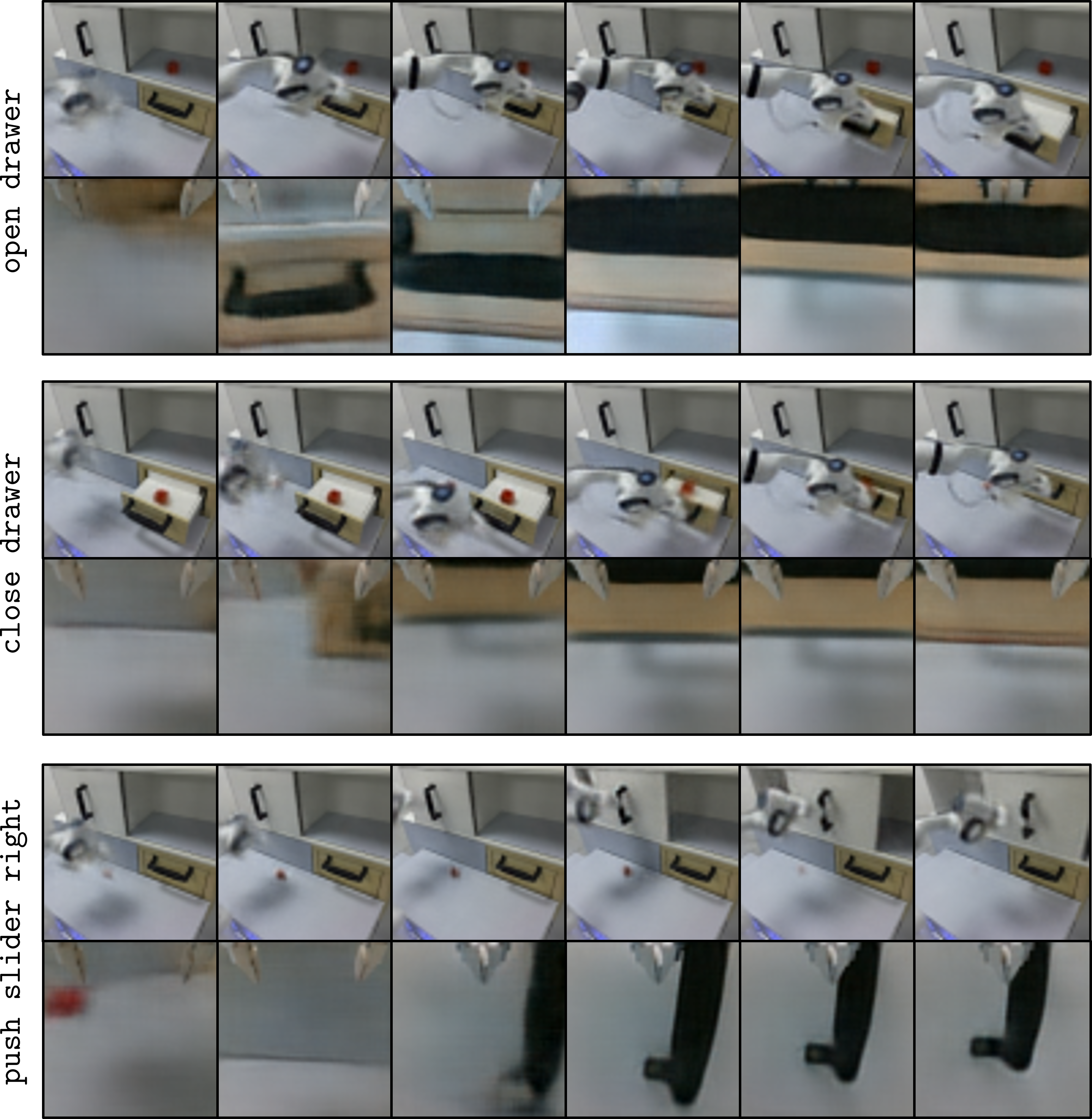}
    \caption{Real-world rollout predictions from the learned world model. Each block shows a segment of a held-out trajectory for a specific skill, with static and gripper camera views decoded from imagined latent states. The model produces accurate long-horizon predictions in real-world settings.}
    \label{fig:app_wm_rollouts}
\end{figure}

We evaluate the predictive capabilities of our learned world model on real-world hold-out trajectories. As illustrated in Figure~\ref{fig:app_wm_rollouts}, the model generates visually coherent and temporally consistent rollouts over extended horizons. To initiate the prediction, we encode the first two frames of an unseen trajectory to establish the initial context. The model then predicts forward for 80 steps in latent space using its recurrent dynamics, despite being trained with sequences of only 50 steps. The decoded reconstructions from the predicted latents reveal that the world model can accurately track key scene elements, such as the robot arm and manipulated objects, even over long horizons. This highlights the model’s ability to learn meaningful dynamics from play data and maintain structured predictions beyond its training horizon.

% \begin{table}[t]
% \footnotesize
% \centering
% \caption{\ourmodel shows improvement when evaluated on four LIBERO-90 scenes. On the other hand, DPPO once again requires hundreds of thousands of online interactions to achieve comparable performance. Results are averaged over three random seeds.}
% \vspace{0.1cm}
% \label{tab:sim_exp_libero}
% \begin{tabular}{cc|c|c}
% \toprule
% \multicolumn{1}{c|}{\multirow{3}{*}{\textbf{Task}}} & \multirow{1}{*}{\textbf{Base}} & \multicolumn{1}{c|}{\textbf{DiWA (Ours)}} & \multicolumn{1}{c}{\textbf{DPPO}} \\ \cmidrule{3-4}
% \multicolumn{1}{c|}{} & \textbf{Diffusion Policy} & \textbf{Offline Fine-Tuning} & \textbf{Online Fine-Tuning} \\ \cmidrule{2-4}
% \multicolumn{1}{c|}{} & \textbf{Success Rate} & \textbf{Success Rate} & \textbf{Env Steps to Match DiWA} \\ \midrule
% \multicolumn{1}{c|}{\liberoodrawer}   & $40.67 \pm 3.06$ & $\textbf{77.33} \pm \textbf{3.06}$ & $000{,}000 \pm 000{,}000$ \\
% \multicolumn{1}{c|}{\liberostove}   & $54.00 \pm 7.21$ & $\textbf{91.33} \pm \textbf{3.08}$ & $000{,}000 \pm 000{,}000$ \\
% \multicolumn{1}{c|}{\liberocbdrawer}   & $27.33 \pm 3.12$ & $\textbf{78.00} \pm \textbf{8.72}$ & $000{,}000 \pm 000{,}000$ \\
% \multicolumn{1}{c|}{\liberoctdrawer}   & $75.33 \pm 2.31$ & $\textbf{100.00} \pm \textbf{0.00}$ & $000{,}000 \pm 000{,}000$ \\
% \midrule
% \multicolumn{2}{c|}{\textbf{Total Physical Interactions:}} & \textbf{0} & $\sim$0.0M \\ \bottomrule
% \end{tabular}
% \vspace{-0.5cm}
% \end{table}

%\input{sections/10-reviews}
%\input{sections/11-rebuttal}
\end{document}